%% file: main.tex
\documentclass[runningheads]{llncs}

\usepackage{eccv}

\usepackage{eccvabbrv}

\usepackage{graphicx}
\usepackage{booktabs}

\usepackage[accsupp]{axessibility}  

\usepackage{hyperref}

\usepackage{orcidlink}

\begin{document}

\title{DRIFT: From Robustness Gaps to Invariance Manifolds for AI-Generated Image Detection} 

\titlerunning{DRIFT}

\author{Abhishek Ameta \and 
Sayan Banerjee \and
Shreyas Pandith \and
Harshit \and
Ankita Chatterjee \and
Akshay Janardan Bankar\orcidlink{0009-0000-3577-1399} \and
Amit Satish Unde}

\authorrunning{A. Ameta, S. Banerjee et al.}

\institute{Samsung Research Institute, Bangalore, India
}

\maketitle

\begin{abstract}
The rapid evolution of generative image models challenges existing AI-generated image detectors, particularly in open-world settings with unseen generators. Recent training-free approaches measure robustness gaps in frozen vision foundation models (VFMs), detecting fakes via perturbation-induced embedding drift. However, these methods rely on fixed invariance geometry inherited from pretraining and lack principled adaptation to the detection task.
We instead formulate AI-generated image detection as learning a structured invariance manifold of real images under one-class supervision. Building upon a frozen VFM, we introduce lightweight projection heads that decompose representation space into complementary robust and fragile subspaces. The robust subspace is explicitly trained to suppress variations induced by physically plausible imaging transformations, approximating tangent directions of a real-image manifold, while the fragile subspace retains sensitivity to edit-like perturbations. A structured ordering margin enforces hierarchical separation between physical invariance and edit-induced variability, enabling detection as a margin-violation test relative to the learned manifold. 
At inference, multi-scale patch-wise drift under both transformation families yields a dual-channel invariance signature and interpretable localization. Extensive experiments demonstrate strong open-world generalization across unseen generators and resolutions, consistently outperforming training-free robustness-based baselines while providing interpretable invariance-violation maps. 
\end{abstract}

\begin{figure}[h]
\centering
\begin{subfigure}{0.45\textwidth}
\centering
\includegraphics[width=\textwidth]{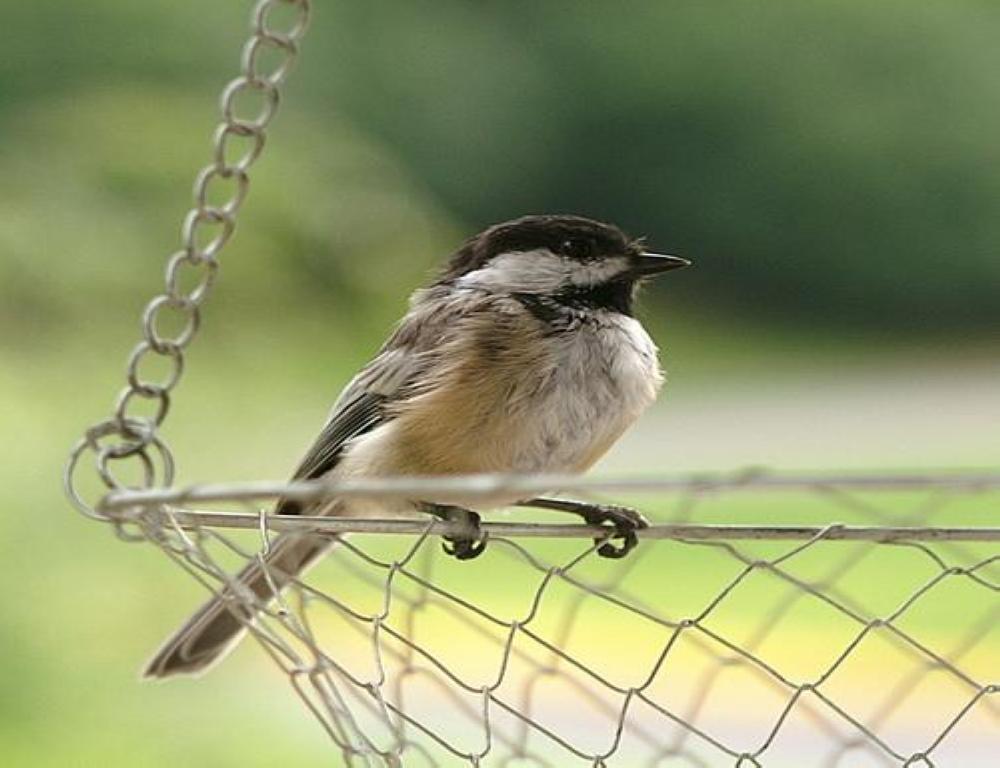}
\caption{Real image.}
\end{subfigure}
\hfill
\begin{subfigure}{0.45\textwidth}
\centering
\includegraphics[width=\textwidth]{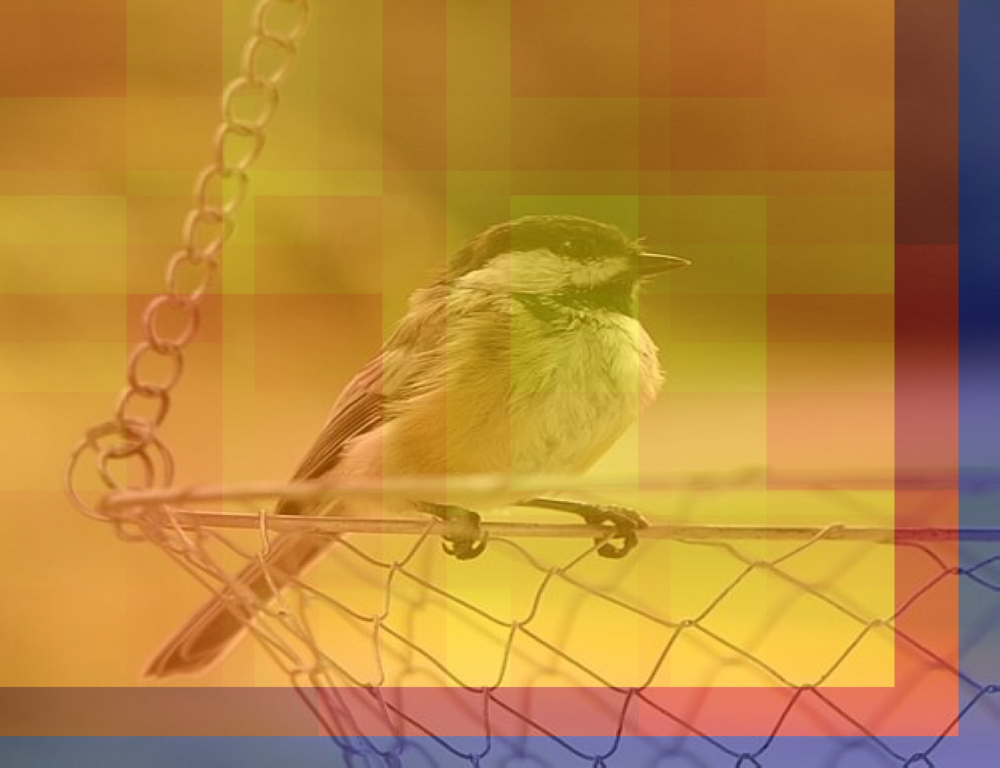}
\caption{Real: Detection heatmap.}
\end{subfigure}
\hfill
\begin{subfigure}{0.45\textwidth}
\centering
\includegraphics[width=\textwidth]{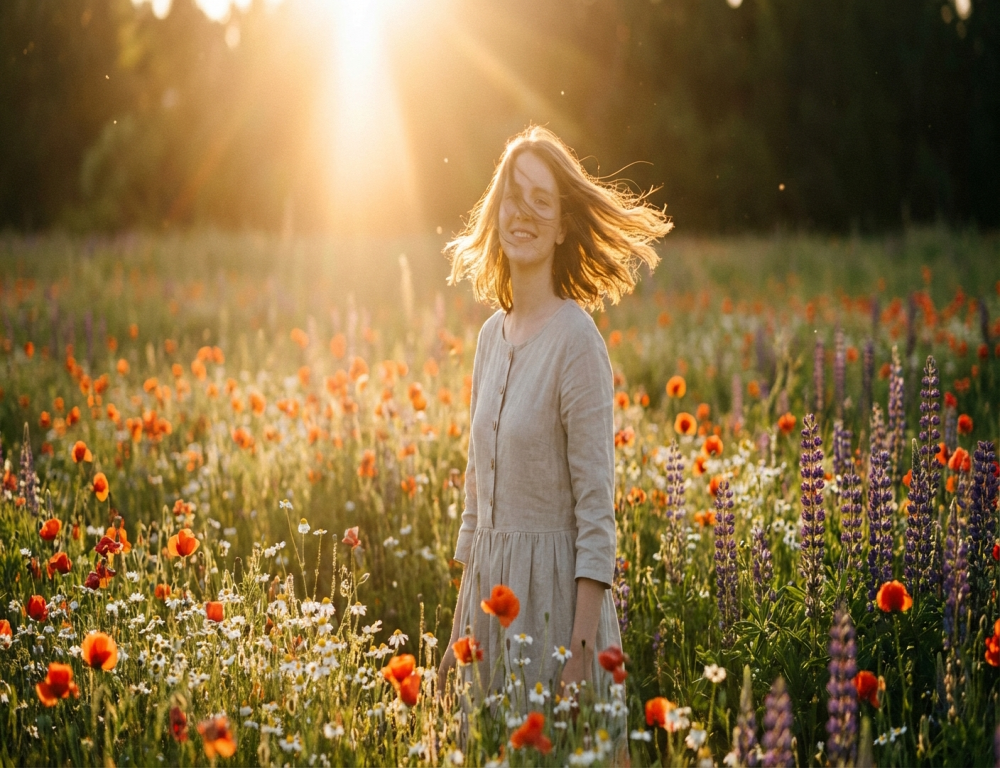}
\caption{AI-generated image.}
\end{subfigure}
\hfill\begin{subfigure}{0.45\textwidth}
\centering
\includegraphics[width=\textwidth]{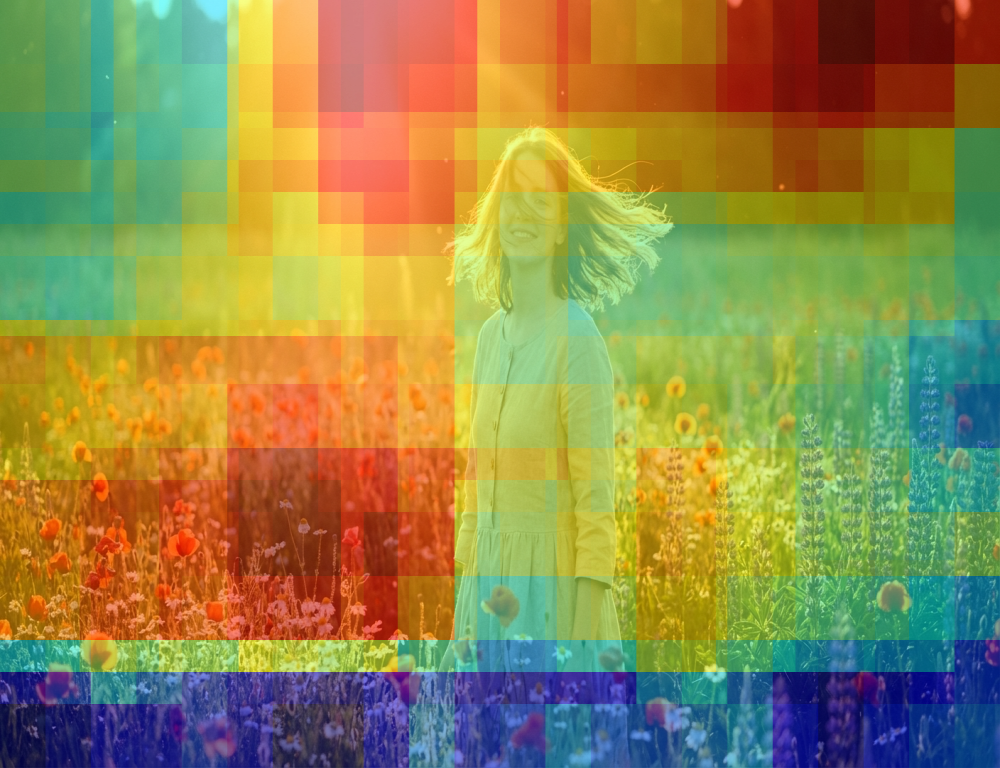}
\caption{AI: Detection heatmap.}
\end{subfigure}
\caption{Visualization of robust-fragile drift violation. Real images maintain drift consistency, whereas AI-generated images show strong drift violations, producing high-response heatmaps.}
\label{fig:heatmap}
\end{figure}

\section{Introduction}
The rapid advancement of generative models has made AI-generated images increasingly photorealistic and difficult to distinguish from real photographs \cite{sd35, sdxl, dalle}. As generation quality improves, detection methods must generalize beyond specific architectures and training paradigms, motivating generator-agnostic detection strategies capable of generalizing across evolving generative paradigms.

Early detection approaches relied primarily on supervised training with labeled real and synthetic data, often exploiting model-specific artifacts or frequency inconsistencies \cite{dire, fire}. While effective in closed-world settings, such methods frequently degrade under unseen generators or distribution shifts. More recent work has explored \emph{training-free} detection strategies using vision foundation models (VFMs) \cite{ssl, vfm}, such as DINOv2 \cite{dinov2}, based on the observation that real and synthetic images exhibit different robustness behavior in representation space \cite{rigid, cropping_robustness, aeroblade}. In particular, prior studies have shown that AI-generated images tend to produce larger embedding drift under certain perturbations (e.g., noise, cropping, resampling), enabling detection via perturb-and-measure scoring \cite{understanding_training_free, rethinking_upsampling}. Although promising, existing methods inherit two fundamental limitations. First, they depend directly on the invariance geometry learned during foundation model pretraining, which is not optimized for authenticity detection. Second, their performance is highly sensitive to the choice of perturbation family, as different corruptions may induce domain-specific biases \cite{rigid, cropping_robustness, rethinking_upsampling}. Consequently, the robustness signal remains passive and fixed, rather than being explicitly shaped for the detection objective. Furthermore, treating robustness as a scalar property overlooks a deeper structural insight: real images lie on a constrained manifold shaped by natural image statistics and physical imaging processes \cite{ferretnet, fatformer, aeroblade}.

In this work, we reformulate AI-generated image detection as a structured invariance learning problem. Rather than measuring drift under a single transformation family, we explicitly decompose representation space into two complementary subspaces: a robust subspace that reduces sensitivity to natural nuisance variations (e.g., mild blur or compression), and a fragile subspace that amplifies sensitivity to edit-like or structurally inconsistent perturbations. We enforce a margin-based ordering constraint between robust and fragile drift magnitudes, thereby learning an invariance hierarchy instead of a scalar robustness score. To stabilize this inequality-constrained optimization under real-only supervision, we introduce an exponential moving average (EMA) \cite{byol, ssl} teacher that provides temporally smoothed geometric targets. EMA acts as a stability operator, decoupling fast parameter updates from invariance ordering evaluation and mitigating collapse or oscillatory margin violations. At inference time, detection is formulated as a geometric margin-violation test. We compute patchwise representation drift under both robust (tangent-like) and fragile (normal-inducing) transformation families, and classify images based on violations of the learned invariance hierarchy between these two subspaces. We project the drift back to image space to produce a heatmap and aggregate scores using a top-$k$ median statistic, yielding both a global authenticity score and interpretable localization. This design bridges global detection and local artifact analysis within a unified invariance-based framework as shown in Fig.~\ref{fig:heatmap}.

Our contributions are summarized as follows:
\begin{itemize}
    \item We propose DRIFT: a real-only, one-class learning framework that explicitly models the physical invariance manifold of real images in representation space.
    \item We introduce a robust–fragile representation factorization with an ordering loss that enforces structured drift geometry under different transform families. Our method separates nuisance physical variations from stronger edit-like variations, reducing sensitivity to perturbation choice and improving calibration in the one-class regime.

    \item We develop a patch-wise invariance drift scoring mechanism that provides both a drift score and interpretable localization, which is not inherent to typical global drift detectors.

    \item Extensive experiments demonstrate that learning structured invariance geometry from real images improves generalization to unseen generators, while maintaining interpretability and robustness in open-world settings.

\end{itemize}

\section{Related Work}
\label{sec:related_work}
Early work focused on identifying artifacts introduced by generative models. CNN-based GAN detectors demonstrated that upsampling artifacts and convolutional inconsistencies can be exploited for detection \cite{cnn_easy_spot, learning_on_gradients}. Subsequent works improved cross-model generalization by explicitly modeling generative artifacts in gradients \cite{learning_on_gradients} or revisiting upsampling operations in CNN generators \cite{rethinking_upsampling}. Universal fake detectors were proposed to improve cross-generator generalization \cite{towards_universal}, and transformation-based training strategies were introduced to enhance robustness to post-processing \cite{improving_synthetic_detection}. While effective in closed-world settings, these supervised methods often rely on model-specific artifacts and degrade under unseen generators or distribution shifts. Some works study synthetic image detection from a transformation perspective, showing that certain image manipulations amplify artifacts and improve generalization \cite{improving_synthetic_detection}. These methods enhance robustness but still operate within supervised detection paradigms. FerretNet models local pixel dependencies and Markovian structure to detect synthetic images efficiently \cite{ferretnet}. By focusing on local statistical inconsistencies, it improves computational efficiency while maintaining generalization. Nonetheless, it remains supervised and artifact-driven.

Frequency-domain inconsistencies have been widely studied. FIRE leverages frequency-guided reconstruction to detect diffusion-generated images \cite{fire}. Spectral learning approaches model real-image spectral distributions and detect deviations without relying on generator-specific artifacts \cite{any_resolution_spectral}. Earlier analyses demonstrated that CNN-generated images exhibit distinctive frequency patterns \cite{cnn_easy_spot}. These methods exploit statistical discrepancies in spectral representations; however, they primarily operate in pixel or frequency space rather than learning structured representation-level invariances.

Reconstruction-based methods assume that generative images exhibit reconstruction inconsistencies under autoencoder or diffusion priors. DIRE measures reconstruction error under diffusion inversion \cite{dire}, while AEROBLADE detects latent diffusion images via autoencoder reconstruction discrepancies \cite{aeroblade}. Other studies analyze intrinsic diffusion sampling characteristics \cite{diffusion_detection}. Although reconstruction-based approaches can capture generative priors, they require generator-specific inversion or autoencoding pipelines and may depend on assumptions about generative architectures. 

Recent works observe that AI-generated images exhibit larger representation drift under perturbations in pretrained VFMs. RIGID measures robustness gaps under noise perturbations in frozen embeddings \cite{rigid}. Follow-up work analyzes the sensitivity of such methods to perturbation choice and backbone robustness \cite{understanding_training_free}. Cropping-based robustness detection leverages invariance inherited from SSL pretraining \cite{cropping_robustness}. These approaches are attractive due to their generator-agnostic and training-free nature. However, they treat representation geometry as fixed and rely on carefully chosen perturbations to induce separable drift. Detection reduces to perturb-and-measure scoring rather than explicitly modeling authentic image variability.

In contrast to supervised artifact modeling, frequency-statistics matching, reconstruction-based inversion, or training-free robustness-gap scoring, we formulate AI-generated image detection as learning a structured invariance manifold of real images under one-class supervision.

\section{Methodology}
\label{sec:method}

\subsection{Problem Setting and Notation}
Let $x \in \mathbb{R}^{H\times W\times 3}$ denote an RGB image.
We assume access to a set of real training images $\mathcal{D}_{\mathrm{real}}=\{x_i\}_{i=1}^N$ and no AI-generated images during training (one-class setting).
At test time, we receive an image $x$ and must predict whether it is real or AI-generated.

Let $\phi:\mathbb{R}^{H\times W\times 3}\rightarrow \mathbb{R}^{d\times m}$ be a frozen VFM feature extractor (e.g., DINOv2), producing $m$ patch tokens each of dimension $d$.
We use two lightweight projection heads:
\begin{align}
h_r(\cdot): \mathbb{R}^{d\times m} \rightarrow \mathbb{R}^{k_r\times m}, \qquad
h_f(\cdot): \mathbb{R}^{d\times m} \rightarrow \mathbb{R}^{k_f\times m},
\end{align}
and define robust and fragile embeddings as:
\begin{align}
z_r(x) = h_r(\phi(x)), \qquad z_f(x) = h_f(\phi(x)).
\end{align}

\begin{figure}[t]
    \centering
    \includegraphics[width=1\linewidth]{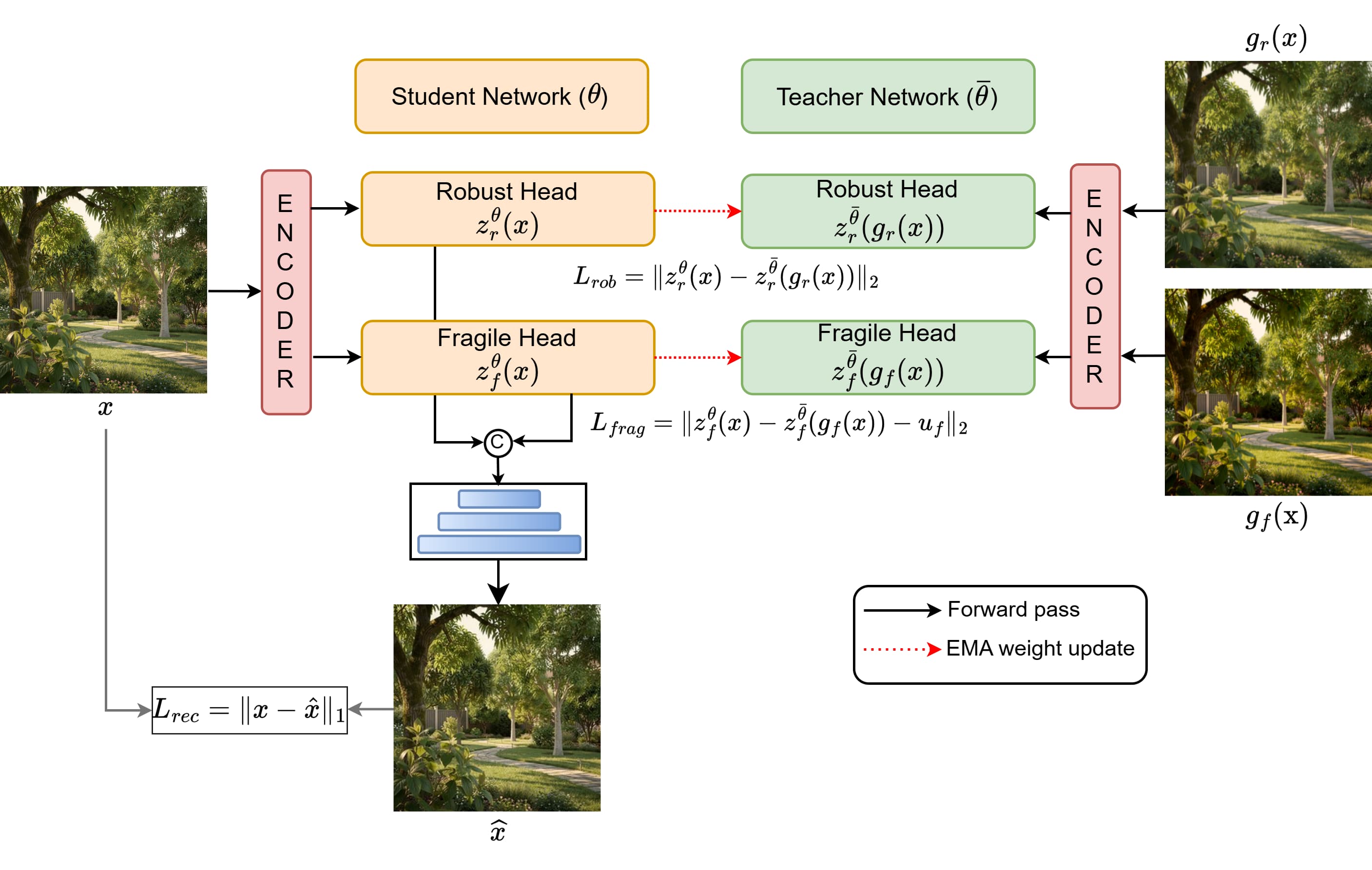}
    \caption{Overview of the proposed student--teacher training framework. 
The student network processes the original image $x$ while the EMA teacher processes transformed views. Two projection heads learn complementary representations. A reconstruction decoder anchors the representation to prevent collapse.}
    \label{fig:main}
\end{figure}

\subsection{Network Architecture and Training Framework}

Fig.~\ref{fig:main} illustrates the overall pipeline of the proposed framework. Given an input image $x$, a shared encoder extracts feature representations which are fed into two projection heads: a robust head and a fragile head. The student network, parameterized by $\theta$, processes the original image $x$ to produce embeddings $z_r^{\theta}(x)$ and $z_f^{\theta}(x)$ corresponding to the robust and fragile branches.

The teacher network, parameterized by $\bar{\theta}$, processes transformed views of the image. A robust transformation $g_r(x)$ generates a nuisance-preserving view, while a fragile transformation $g_f(x)$ produces a structurally perturbing view. These transformed inputs are passed through the teacher encoder and corresponding projection heads to obtain $z_r^{\bar{\theta}}(g_r(x))$ and $z_f^{\bar{\theta}}(g_f(x))$.

The robust branch enforces invariance between the student representation of the original image and the teacher representation of the robustly transformed image through the loss

\[
L_{\text{rob}} = \| z_r^{\theta}(x) - z_r^{\bar{\theta}}(g_r(x)) \|_2 .
\]

The fragile branch captures transformation sensitivity using

\[
L_{\text{frag}} = \| z_f^{\theta}(x) - z_f^{\bar{\theta}}(g_f(x)) - u_f \|_2 ,
\]

where $u_f$ denotes a fragile drift centering term.

To prevent representation collapse, the embeddings are passed through a decoder to reconstruct the input image $\hat{x}$. A reconstruction loss anchors the representation

\[
L_{\text{rec}} = \|x - \hat{x}\|_1 .
\]

The teacher parameters $\bar{\theta}$ are updated using an exponential moving average of the student parameters.

\subsection{Manifold View: Physical Invariance as Geometry}
We model real images as samples from a low-dimensional physical-image manifold $\mathcal{M}_{\mathrm{real}} \subset \mathbb{R}^{H\times W\times 3}$ induced by the camera imaging pipeline and natural scene statistics.
We further define a group (or semigroup) of physically plausible transformations
$\mathcal{G}_R$ (e.g., mild JPEG, mild blur, photometric jitter, resampling round-trips),
and assume approximate closure:
\begin{align}
x\in \mathcal{M}_{\mathrm{real}},\; g\in \mathcal{G}_R \;\Rightarrow\; g(x) \in \mathcal{M}_{\mathrm{real}} \;\;\text{(up to small distortion)}.
\label{eq:closure}
\end{align}

Training-free detectors implicitly rely on invariances already encoded in $\phi$ and score an image via embedding drift $\|\phi(x)-\phi(g(x))\|$ for chosen perturbations.
Our novelty is to learn a detection-adapted invariance geometry from real images only by introducing:
(i) robust/fragile factorization, (ii) a structured ordering objective, and (iii) a stability anchor (EMA teacher + reconstruction), thereby shaping the representation manifold in which invariance violations become more discriminative and better calibrated.

\subsection{Decomposing the Representation Space}
We introduce two transformation families:
\begin{itemize}
    \item \textbf{Robust/Physical transforms} $\mathcal{G}_R$: small camera/codec-like perturbations that preserve semantic content and are common in real acquisition/distribution.
    \item \textbf{Fragile/Edit-like surrogate transforms} $\mathcal{G}_F$: stronger or more ``editing''-like distortions (e.g., strong blur, heavy compression, pixelation, local edits) used only during training to shape a complementary sensitivity direction.
\end{itemize}

\subsection{Drift Measures as Manifold-Violation Scores}
For any embedding $z(\cdot)$ and transform $g$, define patchwise drift:
\begin{align}
\Delta_z(x;g) = z(x) - z(g(x)) \in \mathbb{R}^{k\times m}.
\end{align}
We summarize drift via an $\ell_2$ token norm averaged over patches:
\begin{align}
D_z(x;g) = \frac{1}{m}\sum_{j=1}^{m} \left\| \Delta_z^{(j)}(x;g)\right\|_2.
\label{eq:drift_def}
\end{align}
We also use expectations over sampled transforms:
\begin{align}
D_R(x) = \mathbb{E}_{g\sim \mathcal{G}_R}\left[D_{z_r}(x;g)\right], \qquad
D_F(x) = \mathbb{E}_{g\sim \mathcal{G}_F}\left[D_{z_f}(x;g)\right].
\label{eq:drift_expectations}
\end{align}

\subsection{Learning Objectives: Shaping the Invariance Manifold}
A central difficulty in learning invariance from real-only data is avoiding representation collapse and unstable drift behavior. To stabilize training, we introduce three complementary stabilization mechanisms: (i) EMA teacher for temporally smoothed targets, (ii) auto-associative reconstruction anchor to preserve representational diversity, and (iii) a strcutured robust-fragile ordering constraint to enforce meaningful geometric separation. We detail each component below.

\subsubsection{Collapse Under Real-Only Invariance Learning.}

Consider a real-only invariance objective of the form
\[
\mathcal{L}_{\mathrm{inv}}
=
\mathbb{E}_{x,g}
\| z_\theta(x) - z_\theta(g(x)) \|_2^2,
\]
which encourages embeddings of transformed views to be similar.

This objective admits a degenerate solution: if the network maps all inputs to a constant vector,
\[
z_\theta(x) \equiv c \quad \forall x,
\]
then transformation-induced drift vanishes and
\[
\mathcal{L}_{\mathrm{inv}} = 0.
\]
Although this minimizes the objective, it destroys meaningful representation structure, since all images become indistinguishable. Such collapse is particularly likely under real-only supervision, where no explicit mechanism enforces diversity across different inputs.

To mitigate this instability, we introduce EMA teacher that provides temporally smoothed invariance targets and stabilizes real-only training. 

\subsubsection{EMA Teacher for Stable Invariance Targets.}
Let $\theta$ denote student parameters (heads and decoder) and $\bar{\theta}$ the EMA teacher parameters:
\begin{align}
\bar{\theta}\leftarrow \tau \bar{\theta} + (1-\tau)\theta,
\end{align}
with $\tau\in[0,1)$.
The teacher provides stable targets for drift matching. The student matches temporally smoothed teacher targets while gradients are applied only to the student.
Because the teacher evolves slowly as a low-pass filtered ensemble of past students, collapse cannot occur instantaneously; degenerate solutions must persist over time to propagate through the EMA filter.
This temporal decoupling stabilizes invariance learning and reduces collapse under real-only training.

\subsubsection{Reconstruction Anchor to Prevent Collapse.}
We add a lightweight decoder $d(\cdot)$ that reconstructs $x$ from a combined latent $u(x)$ (e.g., concatenation or fusion of $z_r$ and $z_f$):
\begin{align}
\hat{x} = d\!\left(u(x)\right), \qquad
\mathcal{L}_{\mathrm{rec}} = \|x-\hat{x}\|_1.
\label{eq:rec_loss}
\end{align}
This forces embeddings to retain image information, preventing trivial constant solutions while still allowing invariance constraints.

\subsubsection{Robust Drift Matching (Learned Physical Invariance).}
For $g\sim\mathcal{G}_R$, we enforce invariance in $z_r$ via teacher--student matching:
\begin{align}
\mathcal{L}_{\mathrm{rob}} =
\mathbb{E}_{x\sim \mathcal{D}_{\mathrm{real}}}\;
\mathbb{E}_{g\sim\mathcal{G}_R}
\left[
\left\| z_r^{\theta}(x)- z_r^{\bar{\theta}}(g(x))\right\|_2^2
\right].
\label{eq:rob_loss}
\end{align}
Unlike training-free scoring approaches, our model learns the invariance structure associated with the robust transformation family $\mathcal{G}_R$ directly from real data.

\subsubsection{Fragile Drift Centering (Avoid Degeneracy and Encourage Complementarity).}
A potential degeneracy arises when the fragile branch collapses to a constant embedding, i.e.,
\[
z_f(x) \equiv c_f \quad \forall x,
\]
which leads to vanishing fragile drift
\[
D_F(x)=\|z_f(x)-z_f(g_F(x))\|_2=0.
\]
In this case the fragile branch carries no meaningful transformation sensitivity and fails to play a complementary role to the robust branch. To prevent this, we impose a weak centering constraint on fragile drift to regularize magnitudes for $g\sim\mathcal{G}_F$:
\begin{align}
\mathcal{L}_{\mathrm{frag}} =
\mathbb{E}_{x\sim \mathcal{D}_{\mathrm{real}}}\;
\mathbb{E}_{g\sim\mathcal{G}_F}
\left[
\left(D_{z_f}(x;g)-\mu_f\right)^2
\right],
\label{eq:frag_loss}
\end{align}
where $\mu_f$ is a target drift level (estimated via running average). This centering operation discourages trivial solutions in which fragile drift collapses uniformly across samples and instead encourages non-zero variability in fragile responses. As a result, the fragile branch maintains sensitivity to structurally disruptive transformations, complementing the robust branch which suppresses nuisance-preserving variations.

\subsubsection{Robust--Fragile Ordering Margin (Structured Geometry).}
To enforce complementary behavior between the two branches, we impose an ordering constraint on transformation-induced drift. Let
\[
D_R(x)=\|z_r(x)-z_r(g_R(x))\|_2, \qquad
D_F(x)=\|z_f(x)-z_f(g_F(x))\|_2
\]
denote robust and fragile drift respectively. Robust transformations represent nuisance-preserving variations and should therefore induce smaller drift than fragile transformations, which probe structurally disruptive perturbations. We enforce this relationship using a margin constraint
\[
\mathcal{L}_{\mathrm{ord}} =
\max\big(0,\, D_R(x) + \gamma - D_F(x)\big),
\]
where $\gamma>0$ is a margin. This loss encourages fragile drift to exceed robust drift by at least $\gamma$, preventing trivial solutions in which both drifts collapse or become indistinguishable. As a result, the robust branch learns invariance to nuisance transformations while the fragile branch remains sensitive to structural deviations, producing complementary representation behavior.

\subsubsection{Total Objective}
The full training objective is:
\begin{align}
\mathcal{L} = \lambda_{\mathrm{rec}}\mathcal{L}_{\mathrm{rec}}
+ \lambda_{\mathrm{rob}}\mathcal{L}_{\mathrm{rob}}
+ \lambda_{\mathrm{frag}}\mathcal{L}_{\mathrm{frag}}
+ \lambda_{\mathrm{ord}}\mathcal{L}_{\mathrm{ord}}.
\label{eq:total_loss}
\end{align}

\subsection{Inference: Detection and Localization}
At inference time, only the student network is used. Given a test image $x$, we evaluate transformation-induced drift under both robust and fragile transformations. The student network produces robust and fragile embeddings $z_r(x),  z_f(x)$, respectively, along with embeddings of the transformed views $z_r(g_r(x)), z_f(g_f(x))$.

We compute the corresponding drift magnitudes

\[
D_R(x) = \|z_r(x) - z_r(g_r(x))\|_2,
\qquad
D_F(x) = \|z_f(x) - z_f(g_f(x))\|_2.
\]

Detection is formulated as a margin violation test based on the learned robust--fragile ordering

\[
S(x) = D_R(x) + \gamma - D_F(x),
\]

where $\gamma$ is the ordering margin. Images with $S(x) > 0$ violate the learned invariance hierarchy and are classified as AI-generated, while images satisfying the ordering constraint are classified as real.

\subsubsection{Drift Map.}
Let $\{x_p\}_{p=1}^{P}$ denote image patches.
We define the patchwise drift score as the margin violation between robust and fragile drift:

\[
S(p) = D_R(p) + \gamma - D_F(p),
\]
where $D_R(p)$ and $D_F(p)$ denote robust and fragile drift respectively.
Collecting scores over all patches forms the spatial drift map

\[
\mathbf{S}_{DM} = \{ S(p) \}_{p=1}^{P}.
\]
This map highlights regions that violate the learned invariance ordering and is visualized as a localization heatmap after bilinear interpolation to the original image resolution.

\subsubsection{Robust Global Scoring via Top-$k$ Median Aggregation.}
Patchwise margin violation scores $S(p)$ can be noisy due to stochastic transformations and local texture variability. To obtain a robust image-level score, we aggregate patch scores using a Top-$k$ median rule. Let $\mathbf{S}_{DM}=\{S(p)\}_{p=1}^{P}$ denote the patchwise drift map and let $\mathbf{S}_{\downarrow}$ be the same set sorted in descending order. We select the Top-$k$ most suspicious patches
\[
\mathcal{T}_k = \{ S_{\downarrow}(1), \ldots, S_{\downarrow}(k) \},
\]
and define the global detection score as
\[
S_{\mathrm{global}}(x) = \mathrm{median}\big(\mathcal{T}_k\big).
\]
This aggregation emphasizes localized artifacts by focusing on high-scoring regions, while the median reduces sensitivity to outlier patches, yielding a stable global score for detection.

\section{Experimental Results}
\label{sec:experiments}

\subsection{Dataset Construction}

\subsubsection{Training data.}
We train our model using only real images from the MIT-5K \cite{mit5k}, LSUN \cite{lsun} and RAISE \cite{raise} datasets. These datasets provide diverse photographs captured under varying illumination conditions, resolutions and camera characteristics. In accordance with our one-class formulation, no AI-generated images are used during training.

\subsubsection{Testing Datasets.}
\label{sec:testing_data}
We evaluate our method on the same benchmark test set adopted in FerretNet \cite{ferretnet} to ensure fair comparison. The evaluation benchmark contains real images and AI-generated images using multiple state-of-the-art generative models, including both GAN-based and diffusion-based approaches. Specifically, we test on the following datasets:

\begin{itemize}
    \item \textbf{ForenSynths}:
    A forensic-oriented benchmark containing synthetic images generated by multiple GAN-based architectures: StyleGAN \cite{style}, BigGAN \cite{biggan}, CycleGAN \cite{cyclegan}, GauGAN \cite{gaunet}, and Deepfake \cite{depfake++}. Real images are sourced from six widely-used datasets: LSUN \cite{lsun}, ImageNet \cite{imagenet}, CelebA \cite{celeb}, CelebA-HQ \cite{celebHQ}, COCO \cite{coco}, and FaceForensics++ \cite{depfake++}.

    \item \textbf{Diffusion-6-Class (Diffusion-6cls)}:
    A diffusion-based benchmark comprising synthetic images generated by six distinct diffusion model variants, representing modern large-scale generative systems including DALL-E \cite{dalle}, Guided \cite{guided}, PNDM \cite{pndm}, VQ-Diffusion \cite{vqdiffusion}, Glide \cite{glide}, and LDM \cite{ldm}. 

      \item \textbf{Synthetic Real-World Test Set (PromptWorld-1K)}: To evaluate generalization to modern generative models, we construct a synthetic test set using two widely used generative systems: Gemini and ChatGPT. We generate 1000 images from diverse prompts, covering common real-world scenes such as people, animals, landscapes, objects, and indoor environments. 
    .

\end{itemize}


\begin{table}[t]
\caption{Accuracy and average precision comparisons of different methods on ForenSynth test set. The best, second-best and third-best results are highlighted in red, blue, and green colors, respectively. Refer to Supplemental for detailed analysis.}
\label{tab:forensynth}
\begin{tabular}{|l|l|l|l|l|l|l|}
\hline
Methods            & StyleGAN   & BigGAN     & CycleGAN   & GauGAN     & Deepfake  & Mean      \\ \hline
Ojha \cite{towards_universal}      & 89.0/98.7  & 90.5/99.1  & 87.9/99.8  & 89.9/100.0 & 80.2/90.2 & 89.1/98.3 \\ \hline
FreqNet \cite{freqnet}   & 90.2/99.7  & 90.5/96.0  & 95.8/99.6  & \textcolor{blue}{93.4/98.6}  & 88.9/94.4 & 91.5/98.5 \\ \hline
NPR \cite{rethinking_upsampling}        & 96.3/99.8  & 87.5/94.5  & 95.0/99.5  & 86.6/88.8  & 77.4/86.2 & 92.5/96.1 \\ \hline
FatFormer \cite{fatformer} & 97.2/99.8  & \textcolor{red}{99.5/100.0} & \textcolor{red}{99.3/100.0} & \textcolor{red}{99.4/100.0} & \textcolor{red}{93.2/98.0} & \textcolor{red}{98.4/99.7} \\ \hline
SAFE  \cite{improving_synthetic_detection}     & \textcolor{green}{98.0/99.9}  & 89.7/95.9  & \textcolor{green}{98.9/99.8}  & 91.5/97.2  & \textcolor{blue}{93.1/97.5} & \textcolor{green}{96.2/98.8} \\ \hline
CO-SPY \cite{cospy}     & 63.9/70.2  & 71.6/83.9  & 58.5/55.8  & 69.6/83.4  & 65.7/79.7 & 65.7/76.1 \\ \hline
FerretNet \cite{ferretnet}          & \textcolor{blue}{98.0/100.0} & \textcolor{green}{92.6/98.5}  & 98.8/99.9  & 91.4/99.8  & 89.2/96.7 & 95.9/99.3 \\ \hline
DRIFT (Ours)   &  \textcolor{red}{98.6/100.0}  & \textcolor{blue}{94.3/99.3}  & \textcolor{blue}{99.1/100}  & \textcolor{green}{92.8/100}  & \textcolor{green}{91.2/97.8} & \textcolor{blue}{97.8/99.8} \\ \hline
\end{tabular}
\end{table}

\begin{table}[t]
\caption{Accuracy and average precision comparisons  of different methods on Diffusion-6-cls test  dataset. The best, second-best and third-best results are highlighted in red, blue, and green colors, respectively. Refer to Supplemental for detailed analysis.}
\label{tab:dalle}
\begin{tabular}{|l|l|l|l|l|l|}
\hline
Dataset      & RIGID \cite{rigid}  & FatFormer \cite{fatformer} & SAFE \cite{improving_synthetic_detection} & FerretNet \cite{ferretnet} & DRIFT (Ours)       \\ \hline
Dall-E       & 43.5/48.0                  & \textcolor{red}{98.8/99.8}         & \textcolor{blue}{97.5/99.7}    & 91.4/98.2         &  \textcolor{green}{92.1/99.2}          \\ \hline
Guided       &  58.0/94.9              & 76.1/92.0         & \textcolor{green}{82.4/95.8}    & \textcolor{blue}{92.1/98.6}         &  \textcolor{red}{92.9/98.9}          \\ \hline
PNDM         & 53.9/90.6                            & \textcolor{red}{99.3/100.0}        & 78.9/98.6    & \textcolor{blue}{96.9/100.0}        &  \textcolor{green}{94.3/100}          \\ \hline
VQ-Diffusion &  53.6/90.9                           & \textcolor{green}{98.0/100.0}       & \textcolor{red}{100.0/100.0}  & \textcolor{blue}{99.9/100.0}        &  97.0/100.0          \\ \hline
Glide-50-27  &   83.8/83.1                          & 94.7/99.4         & \textcolor{green}{96.6/99.2}    & \textcolor{red}{97.2/99.7}         &   \textcolor{blue}{96.8/99.3}         \\ \hline
LDM-100      &   53.8/58.6                          & \textcolor{blue}{98.7/99.9}         & \textcolor{red}{98.8/100.0}   & 95/100.0        & \textcolor{green}{98.6/100.0} \\ \hline

\end{tabular}
\end{table}

\subsection{Implementation and Optimization Details}
\label{sec:impl_details}

\textbf{Backbone and heads.}
We use a frozen DINOv2 ViT-B/14 backbone to extract a 768-d global image representation. On top of the backbone, we attach two lightweight projection heads (robust head $z_r$ and fragile head $z_f$). Each head is a 2-layer MLP with architecture $768\!\rightarrow\!512\!\rightarrow\!256$ using GELU activation, with LayerNorm applied before the first linear layer. The output embedding dimension is $d{=}256$. Only the projection heads are trainable.

\textbf{Hyperparameters.}
We optimize a weighted sum of robust invariance, fragile sensitivity, and ordering margin losses, with ordering margin $\gamma{=}0.3$ and weights $(\lambda_{rob},\lambda_{frag},\lambda_{ord})=(1.0,1.0,0.5)$. We additionally use a reconstruction anchor regularizer with weight $0.1$ to avoid degenerate collapse.

\textbf{EMA teacher.}
To stabilize one-class training, we maintain an exponential moving average (EMA) teacher for the projection heads. Teacher parameters are updated as $\theta_{\text{ema}}\leftarrow \lambda_{\text{ema}}\theta_{\text{ema}}+(1-\lambda_{\text{ema}})\theta$ with momentum $\lambda_{\text{ema}}{=}0.996$, and are used to compute stable targets during drift estimation.

\textbf{Optimization.}
We train with AdamW using learning rate $3{\times}10^{-4}$, weight decay $0.01$, batch size $64$, for $50$ epochs. We apply a cosine learning-rate schedule with linear warmup for the first $5$ epochs and clip gradients to a maximum norm of $1.0$. All images are resized to $224{\times}224$ during training and evaluation.

\textbf{Performance Evaluation.} Following previous work \cite{ferretnet, rigid}, Accuracy (ACC) and Average Precision (AP) are used as the primary evaluation metrics.

\subsection{Main Results}

\subsubsection{Results on GAN-based generators (ForenSynths).}
On the ForenSynth benchmark, our method achieves consistently strong performance across all generator families as shown in Table~\ref{tab:forensynth}. These results indicate that explicitly learning the invariance manifold of real images leads to a more stable detection signal than relying solely on pretrained feature geometry.
\vspace{-15pt}

\subsubsection{Results on diffusion-based generators (Diffusion-6cls).}
We further evaluate in Table~\ref{tab:dalle} the performance of the proposed method on the Diffusion-6cls benchmark. Compared with training-free detectors such as RIGID, our method provides more stable performance across different diffusion sampling strategies. This demonstrates that the learned invariance manifold captures structural properties of real images that generalize beyond specific generator architectures.

\vspace{-15pt}
\subsubsection{Results on synthetic real-world test set (PromptWorld-1K).}
As shown in Table~\ref{tab:gemini}, our method achieves $93.2\%$ ACC / $92.0\%$ AP on Gemini and $94.8\%$ ACC / $95.0\%$ AP on ChatGPT, outperforming training-free robustness detectors such as RIGID and frequency-based approaches such as FIRE. These results demonstrate that learning a structured invariance manifold from real images enables strong generalization to unseen prompt-based generators

\begin{table}[t]
\caption{Accuracy and average precision comparisons of different methods on Synthetic real-world test set (PromptWorld-1K). The best, second-best and third-best results are highlighted in red, blue, and green colors, respectively.}
\label{tab:gemini}
\begin{tabular}{|l|l|l|l|l|l|}
\hline
              & FIRE \cite{fire}                                                                                    & RIGID \cite{rigid}                              & FerretNet \cite{ferretnet}                                                                              & NPR \cite{rethinking_upsampling}                                                                                   & DRIFT (Ours)                                                                          \\ \hline
Ref           & CVPR`25                                                                                 & Arxiv`25                           & NeurIPS`25                                                                             & CVPR`24                                                                                & \multicolumn{1}{c|}{-}                                                        \\ \hline
Training data & \multicolumn{1}{c|}{\begin{tabular}[c]{@{}c@{}}Labelled \\ real/fake data\end{tabular}} & \multicolumn{1}{c|}{Training-free} & \multicolumn{1}{c|}{\begin{tabular}[c]{@{}c@{}}Labelled\\ real/fake data\end{tabular}} & \multicolumn{1}{c|}{\begin{tabular}[c]{@{}c@{}}Labelled\\ real/fake data\end{tabular}} & \multicolumn{1}{c|}{\begin{tabular}[c]{@{}c@{}}Only real\\ data\end{tabular}} \\ \hline
Gemini   &     \textcolor{green}{78.3/61.2}                                                                                   &  40.3/30.5                           & 50.9/66                                                                              & \textcolor{blue}{91.9/95.2}                                                                              & \textcolor{red}{93.2/92}                                                                     \\ \hline
ChatGPT       &    77.04/56.2                                                                                     & 41.3/34                         & \textcolor{red}{99.09/99.1}                                                                             & \textcolor{blue}{95.5/98.3}                                                                              & \textcolor{green}{94.8/95}                                                                     \\ \hline
\end{tabular}
\end{table}

\begin{figure}[h]
\centering
\begin{subfigure}{0.45\textwidth}
\centering
\includegraphics[width=\textwidth]{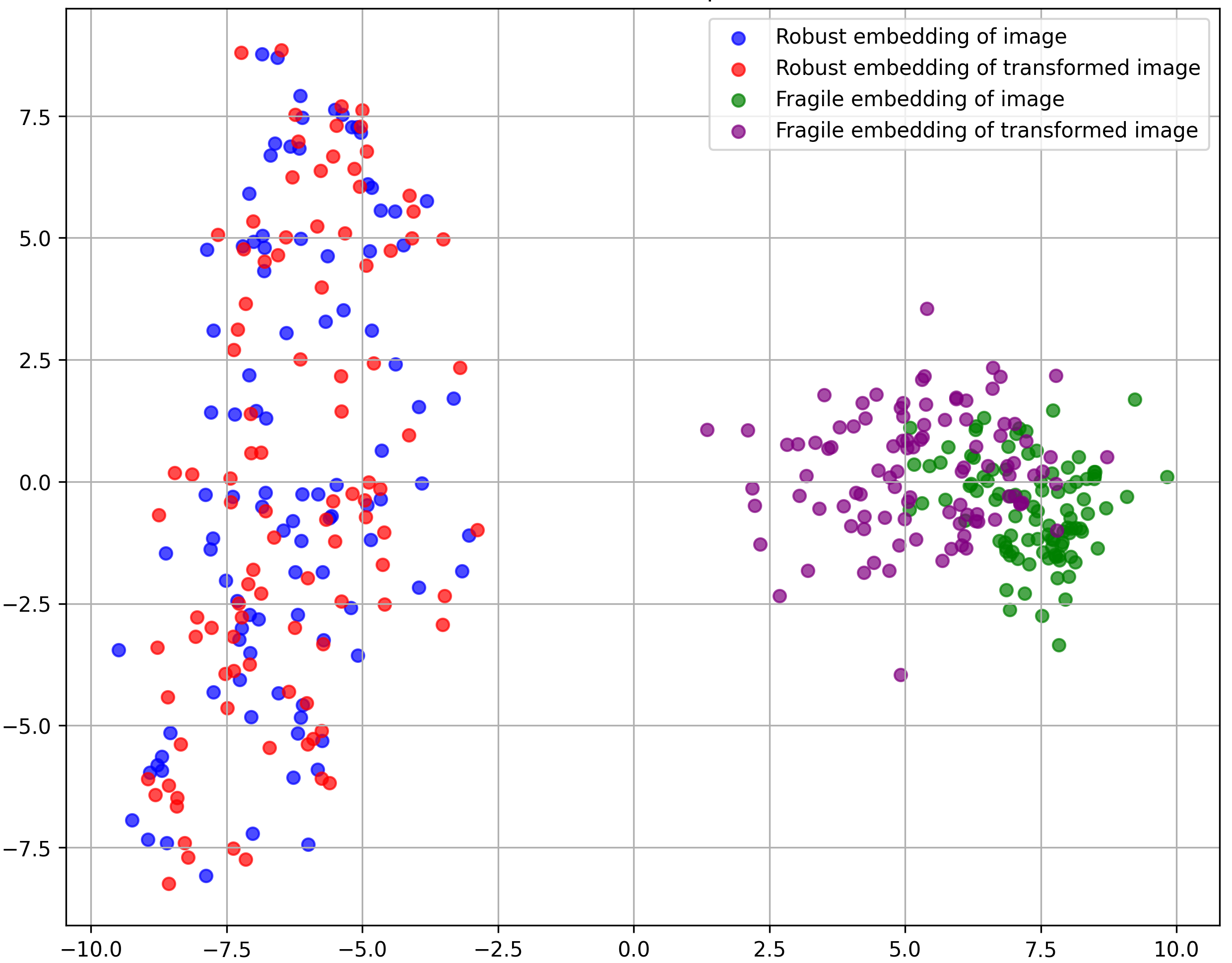}
\caption{t-SNE of embeddings under robust and fragile transformations.}
\end{subfigure}
\hfill
\begin{subfigure}{0.45\textwidth}
\centering
\includegraphics[width=\textwidth]{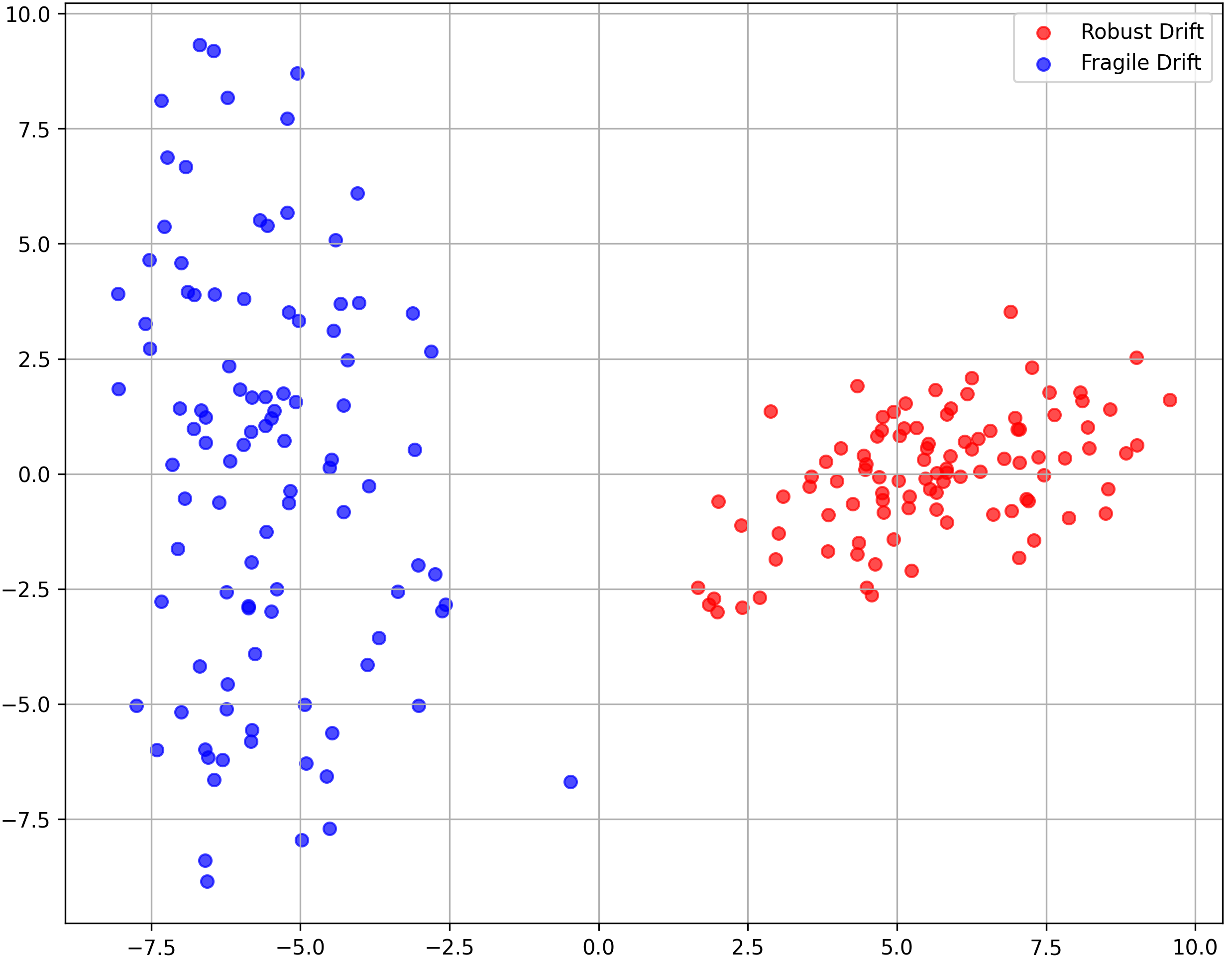}
\caption{t-SNE of robust vs fragile embedding drift.}
\end{subfigure}
\caption{Representation geometry under robust and fragile transformations.}
\label{fig:tsne}
\end{figure}

\subsection{Ablation Studies}
We perform ablations in Table~\ref{tab:ablation} to analyze the contribution of each component of the proposed framework.
\vspace{-15pt}
\subsubsection{Robust vs.\ Fragile Heads.}
We compare the proposed dual-head design with three alternatives: robust-only, fragile-only, and a shared single-head representation. The robust-only model achieves reasonable performance, while the fragile-only variant performs significantly worse. The dual-head model consistently achieves the best results, confirming that robust invariance and fragile sensitivity provide complementary signals for detecting AI-generated images.
\vspace{-15pt}
\subsubsection{Training Stability Mechanisms.}
We evaluate the importance of the EMA teacher, reconstruction anchor, and ordering margin. Removing the EMA teacher leads to reduced accuracy due to unstable representation targets. Removing the reconstruction anchor causes a larger degradation, indicating that reconstruction regularization helps prevent representation collapse when training only on real images. 
\vspace{-15pt}
\subsubsection{Transformation Sensitivity.}
We analyze different robust and fragile perturbation families. Robust transformations such as mild blur and JPEG compression produce stable detection performance, while fragile perturbations such as pixelation, resampling, and defocus blur consistently reveal structural inconsistencies in generated images. A leave-one-out analysis confirms that the detector does not rely on any single perturbation family.

\begin{table}[!]
	\caption{Ablation results on the SDXL-Turbo dataset}
    \label{tab:ablation}
	\centering
	\begin{subtable}[t]{0.48\textwidth}
		\begin{tabular}{lccc}
\toprule
\textbf{Variant} & \textbf{AUC} $\uparrow$ & \textbf{AP} $\uparrow$  \\
\midrule
Robust-only & 96.3 & 98.4  \\
Fragile-only & 61.2 & 78.3   \\
Shared single head & 94.6   & 97.3 \\
Dual-head  & 98.1  & 100.0   \\
\bottomrule
\end{tabular}
		\caption{Robust vs Fragile decomposition}
		\label{tab:ab_1}
	\end{subtable}
	\hspace{\fill}
	\begin{subtable}[t]{0.48\textwidth}	
		\begin{tabular}{lccc}
\toprule
\textbf{Variant} & \textbf{AUC} $\uparrow$ & \textbf{AP} $\uparrow$  \\
\midrule
Full model & 98.1 & 100.0   \\
-- EMA teacher & 94.3  & 97.5  \\
-- Reconstruction anchor & 81.3 & 84.2  \\
-- Ordering margin & 96.7  & 95.3   \\
\bottomrule
\end{tabular}
		\caption{Network architecture}
		\label{tab:ab_2}
	\end{subtable}

	\begin{subtable}[t]{0.48\textwidth}
		\begin{tabular}{lccc}
\toprule
\textbf{Robust family} & \textbf{AUC} $\uparrow$ & \textbf{AP} $\uparrow$ \\
\midrule
JPEG & 85.7  & 88.4   \\
Mild blur & 85.3 & 90.2   \\
Resampling & 87.3 & 93.5   \\
Photometric jitter & 61.3 & 79.2   \\
\bottomrule
\end{tabular}
		\caption{Robust perturbation family sensitivity}
		\label{tab:ab_3}
	\end{subtable}
	\hspace{\fill}
	\begin{subtable}[t]{0.49\textwidth}
		
\begin{tabular}{lccc}
\toprule
\textbf{Fragile family} & \textbf{AUC} $\uparrow$ & \textbf{AP} $\uparrow$  \\
\midrule
Pixelation & 66.3 & 79.3   \\
Defocus blur & 63.1 & 80.3   \\
Resampling & 68.7  & 78.3   \\
Photometric jitter & 59.3 & 76.4.   \\
\bottomrule
\end{tabular}
		\caption{Fragile perturbation family sensitivity}
		\label{tab:ab_4}
	\end{subtable}\
    \label{tab:table1}
\end{table}

\subsection{Visualization of representation geometry.}
Fig.~\ref{fig:tsne} visualizes the embedding behavior under robust and fragile transformations using t-SNE projections. As shown in Fig.~\ref{fig:tsne}(a), embeddings of the original images and their robustly transformed versions lie close in the representation space, indicating that the robust branch successfully learns invariance to nuisance-preserving transformations. In contrast, embeddings under fragile transformations exhibit larger dispersion, as shown in Fig.~\ref{fig:tsne}(b). This confirms that the fragile branch remains sensitive to structurally disruptive perturbations, producing larger representation drift. These observations validate the complementary roles of the robust and fragile subspaces in modeling the invariance structure of real images.

\section{Conclusions}
In this work, we reformulated AI-generated image detection as a structured invariance learning problem. Instead of relying on fixed robustness gaps in pretrained representations, we explicitly learned the physical invariance manifold of real images using only real data. By decomposing the representation space into complementary robust and fragile subspaces and enforcing an ordering constraint between their transformation-induced drifts, the proposed framework captures the geometric structure of natural images and detects violations introduced by generative models.
Extensive experiments across GAN-based, diffusion-based, and prompt-driven generators demonstrate that the learned invariance geometry provides strong open-world generalization while maintaining interpretable localization through patch-wise drift maps. The results indicate that modeling invariance structure rather than generator-specific artifacts offers a promising direction for robust AI-generated image detection. 

\bibliographystyle{splncs04}
\bibliography{main}

\input{sec/supplementary}

\end{document}

%% file: sec/supplementary.tex
 \clearpage
\setcounter{page}{1}
\maketitlesupplementary

\section{Geometric Interpretation of Robust and Fragile Transformations}
\label{sec:geometry}

We provide a geometric interpretation of robust and fragile transformations in terms of tangent and normal directions of the real-image manifold.

\paragraph{Real-image manifold.}
Let $\mathcal{X} \subset \mathbb{R}^{n}$ denote the ambient image space.
Natural images concentrate on a low-dimensional manifold:

\[
\mathcal{M} \subset \mathcal{X}, \qquad \dim(\mathcal{M}) \ll n.
\]

For each $x \in \mathcal{M}$, the ambient space admits a local orthogonal decomposition:

\[
\mathbb{R}^{n} = T_x\mathcal{M} \oplus N_x\mathcal{M},
\]

where $T_x\mathcal{M}$ is the tangent space and $N_x\mathcal{M}$ is the normal space.

\paragraph{Tangent directions induced by robust transforms.}
Consider a smooth transformation family $g_R(x;\alpha)$ satisfying:

\[
g_R(x;0) = x,
\]

and for sufficiently small $\alpha$,

\[
g_R(x;\alpha) \in \mathcal{M}.
\]

Then its infinitesimal variation

\[
v_R = \left.\frac{d}{d\alpha} g_R(x;\alpha)\right|_{\alpha=0}
\]

lies in the tangent space:

\[
v_R \in T_x\mathcal{M}.
\]

Robust transformations such as mild blur, JPEG compression, resampling, and photometric adjustments simulate physically plausible acquisition variability.
For small severity, these operations preserve natural image statistics and therefore approximate motion along the manifold.
Consequently, robust perturbations predominantly induce tangent directions.

\paragraph{Normal components induced by fragile transforms.}
In contrast, consider a transformation family $g_F(x;\alpha)$ for which small perturbations violate manifold constraints:

\[
g_F(x;\alpha) \notin \mathcal{M}
\quad \text{for small } \alpha.
\]

Its first-order variation

\[
v_F = \left.\frac{d}{d\alpha} g_F(x;\alpha)\right|_{\alpha=0}
\]

generally decomposes as

\[
v_F = v_T + v_N,
\quad
v_T \in T_x\mathcal{M}, \;
v_N \in N_x\mathcal{M}.
\]

For edit-like perturbations (pixelation, aggressive resampling, strong defocus blur, strong photometric distortion), the normal component dominates:

\[
\|v_N\| \gg \|v_T\|.
\]

Thus fragile transformations approximate directions with significant normal components, corresponding to off-manifold deviations.

\subsection{Connection to Representation Drift}

For a small perturbation \( g(x) = x + \delta x \), first-order Taylor expansion gives

\[
z_r(g(x))
\approx
z_r(x)
+
J_{z_r}(x)\,\delta x,
\]

where \( J_{z_r}(x) \in \mathbb{R}^{d \times n} \) is the Jacobian.

Thus drift magnitude satisfies

\[
\| z_r(g(x)) - z_r(x) \|
\approx
\| J_{z_r}(x)\,\delta x \|.
\]

If \( \delta x = v_R(x) \in T_x\mathcal{M} \), training enforces

\[
J_{z_r}(x)\, v_R(x) \approx 0,
\]

which suppresses tangent directions (robust invariance).

If \( \delta x \) contains strong normal components (fragile transforms), then

\[
\| J_{z_r}(x)\,\delta x \|
\]
remains large.

Thus, drift magnitude acts as a proxy for deviation from the learned manifold.

\section{Proof of Representation Drift}
\label{sec:proof}
Let x is an image belongs to the image space $\mathcal{X} \in \mathbb{R}^n$.
Let the feature projection, $z:\mathcal{X}\to\mathcal{M}$ project an image to a smooth $k$ ($k<<n$) dimensional Riemannian manifold $(\mathcal{M}, g)$, where $g$ is the Riemannian metric tensor. We have two classes of transforms, one class is the robust/physical transforms denoted as $\mathcal{G}_r$ and another class is the fragile/edit-like surrogate transform denoted as $\mathcal{G}_f$.
The robust transformation is a transformation that does not change semantic information of the input image (the $L_p$ norm of the perturbation is bounded). We can express the robust transformation as a linear path $\gamma_r(t) = x + t\delta x$ in the image space, where $\delta x$ is the direction of the perturbation with $t$ is the scale ($t\to 0$). This transformation creates a corresponding feature transformation $g_r(x + t\delta x)\in\mathcal{M}$, a curve $\gamma_{g_r}(t)$ in the manifold where $g_r\in\mathcal{G}_r$. Hence, the length of the curve $L(\gamma_{g_r})$ signifies the feature drift denoted as $dz_{g_r}(\delta x)$ caused by the transformation $g_r(\cdot)$:
\begin{equation} L(\gamma_{g_r})=dz_{g_r}(\delta x)= \int_{0}^t\Big\lvert\Big\rvert\frac{d\gamma_{g_r}(\tau)}{d\tau}\Big\lvert\Big\rvert_gd\tau=\int_{0}^t\Big\lvert\Big\rvert\frac{\partial{g_r(x + \tau\delta x)}}{\partial \tau}\Big\lvert\Big\rvert_gd\tau
\label{eqn:arc_length}
\end{equation}
For a sufficiently small $t$, the feature drift can be further simplified to,
\begin{equation} 
dz_{g_r}(\delta x)= t\lvert\rvert J_{g_r}(x)\delta x\lvert\rvert_g
\end{equation}
where $J_{g_r}(x)\in\mathbb{R}^{k\times n}$ is the Jacobian of the image feature $z(x)$ with respect to the image $x$. For sufficiently small $t$, the feature drift $dz_{g_r}(\delta x)$ is the linear map that takes your perturbation vector $\delta x$ in the image space and translates it into the corresponding velocity vector in the feature manifold's tangent space, $T_{z(x)}\mathcal{M}$. It measures the feature sensitivity of the neural network with respect to image pixels. It tells us exactly which pixel-level changes, the network cares about and which ones it ignores.

The Jacobian, $J_{g_r}(x)$ can be further decomposed using SVD, resulting the following equation,
\begin{equation}
    dz_{g_r}(\delta x) = t\lvert\rvert U\Sigma V^T\delta x\lvert\rvert_g =t\Big\lvert\Big\rvert\sum_{i=1}^ku_i\sigma_iv_i^T\delta(x)\Big\lvert\Big\rvert_g
    \label{eqn:SVD}
\end{equation}
The columns of $U\in\mathbb{R}^{k\times k}$ (called left singular vectors, $u_i\in\mathbb{R}^n$) form an orthonormal basis for the directions of movement in the feature manifold. The columns of $V\in\mathbb{R}^{n\times n}$ (called right singular vectors, $v_i$) form an orthonormal basis for every possible way you can perturb the image pixels. $\Sigma \in \mathbb{R}^{k\times n}$ is a diagonal matrix consists of singular values $\sigma_1\geq \sigma_2\geq...\geq \sigma_k$ as diagonal elements. These singular values represent how much a specific pixel perturbation is amplified into a semantic shift. 
Because $g_r(\cdot)$ is a semantic-preserving transformation, it aligns almost entirely with the right singular vectors associated with the smallest, near-zero singular values in the tail of $\Sigma$. Let $\sigma_{g_r}$ be the maximum singular value, specific to a particular robust transform $g_r\in\mathcal{G}_r$,  among all of this smallest near-zero singular values. Then $d_{z_r}(\delta x)$ can be bounded as,
\begin{equation}
    dz_{g_r}(\delta x)\leq t\sigma_{g_r}\lvert\rvert\delta x\lvert\rvert_g
\end{equation}
We can express any fragile transformation $g_f\in\mathcal{G}_f$ as a non-semantic preserving path $\gamma_f(t)$ in the image space. This transformation creates a corresponding feature transformation, $g_f(\gamma_f(t))\in\mathcal{M}$, a curve $\gamma_{g_f}(t)$ in the manifold. The feature drift denoted as $dz_{g_f}(\gamma_f(t))$ of the transformation $g_f(\cdot)$ can be expressed as:
\begin{equation} dz_{g_f}(\gamma_f(t))= \int_{0}^t\Big\lvert\Big\rvert\frac{d\gamma_{g_f}(\tau)}{d\tau}\Big\lvert\Big\rvert_gd\tau
\label{eqn:arc_length_fragile}
\end{equation}
Here $\frac{d\gamma_{g_f}(\tau)}{d\tau}$ is the direction of the feature drift, a vector field defined at the tangent space $T_{\gamma_{g_f(t)}}\mathcal{M}$. This vector field  can be represented as, 
\begin{equation}
    \frac{d\gamma_{g_f}(\tau)}{d\tau} = J_{g_f}(\gamma_f(\tau))\frac{d\gamma_f(\tau)}{d\tau}
\end{equation}
Resulting the following expression of the feature drift,
\begin{equation}
    dz_{g_f}(\gamma_f(t))=\int_{0}^t\Big\langle J_{g_f}(\gamma_f(\tau))\frac{d\gamma_f(\tau)}{d\tau}, J_{g_f}(\gamma_f(\tau))\frac{d\gamma_f(\tau)}{d\tau}\Big\rangle_{\gamma_{g_f(\tau)}}d\tau
\end{equation}
where $\langle\cdot,\cdot\rangle_{\gamma_{g_f(\tau)}}$ is the Riemannian inner product defined in the tangent space $T_{\gamma_{g_f(\tau)}}\mathcal{M}$. And $J_{g_f}\in\mathbb{R}^{k\times n}$ is the Jacobian of the transformation $g_f(\cdot)$. 
For fragile transformation, as it incur semantic edit, the drift in the image space $\frac{d\gamma_f(\tau)}{d\tau}$ aligns with the right singular vectors with the largest singular values in the head of the singular matrix obtained by the SVD of $J_{g_f}$.

Let $\sigma_{g_f}$ is the lowest singular value among all of those largest singular values pertaining to the transformation $g_f$. Then the feature drift corresponding to the transformation can be represented as,
\begin{equation}
    dz_{g_f}(\gamma_f(t))\geq\int_0^t\sigma_{g_f}(\tau)\lvert\rvert\dot{\gamma}_f(\tau)\lvert\rvert_g d\tau
\end{equation}
Conclusion: As $\sigma_{g_f}>>\sigma_{g_r},\forall g_f\in\mathcal{G}_f \hspace{0.5cm}\text{and}\hspace{0.5cm}g_r\in\mathcal{G}_r$, the expected feature drift for all of the class of fragile transformation, $\mathbb{E}_{g_f\sim\mathcal{G}_f}d_{z_{g_f}}(\gamma_f(t))$ is much much greater than  the expected feature drift for all of the class of robust transformation, $\mathbb{E}_{g_r\sim\mathcal{G}_r}d_{z_{g_r}}(\delta x)$.

\begin{table}[]
\caption{Accuracy and average precision comparisons of different methods on ForenSynth test set. The best, second-best and third-best results are highlighted in red, blue, and green colors, respectively.}
\label{table:1}
\begin{tabular}{|l|l|l|l|l|l|l|}
\hline
Methods            & StyleGAN    & BigGAN      & CycleGAN    & GauGAN      & Deepfake  & Mean      \\ \hline
Wang \cite{cnn_easy_spot}       & 63.8/91.4  & 52.9/73.3  & 72.7/88.6  & 63.9/92.2  & 51.7/62.3 & 67.1/86.9 \\ \hline
F3Net \cite{f3net}     & 92.6/99.7  & 65.3/69.9  & 76.4/84.3  & 58.1/56.7  & 63.5/78.8 & 80.4/86.2 \\ \hline
FrePGAN \cite{frepgan}  & 80.7/89.6  & 69.2/71.1  & 71.1/74.4  & 60.3/71.7  & 70.9/91.9 & 79.4/87.2 \\ \hline
BiHPF \cite{bihpf}     & 76.9/75.1  & 84.9/81.7  & 81.9/78.9  & 69.5/78.1  & 54.4/54.6 & 78.6/77.9 \\ \hline
LGrad \cite{learning_on_gradients}     & 94.8/99.9  & 82.9/90.7  & 85.3/94.0  & 72.4/79.3  & 58.0/67.9 & 86.1/91.5 \\ \hline
Ojha \cite{towards_universal}      & 89.0/98.7  & 90.5/99.1  & 87.9/99.8  & 89.9/100.0 & 80.2/90.2 & 89.1/98.3 \\ \hline
FreqNet \cite{freqnet}   & 90.2/99.7  & 90.5/96.0  & 95.8/99.6  & \textcolor{blue}{93.4/98.6}  & 88.9/94.4 & 91.5/98.5 \\ \hline
NPR \cite{rethinking_upsampling}        & 96.3/99.8  & 87.5/94.5  & 95.0/99.5  & 86.6/88.8  & 77.4/86.2 & 92.5/96.1 \\ \hline
FatFormer \cite{fatformer} & 97.2/99.8  & \textcolor{red}{99.5/100.0} & \textcolor{red}{99.3/100.0} & \textcolor{red}{99.4/100.0} & \textcolor{red}{93.2/98.0} & \textcolor{red}{98.4/99.7} \\ \hline
SAFE  \cite{improving_synthetic_detection}     & \textcolor{green}{98.0/99.9}  & 89.7/95.9  & \textcolor{green}{98.9/99.8}  & 91.5/97.2  & \textcolor{blue}{93.1/97.5} & \textcolor{green}{96.2/98.8} \\ \hline
CO-SPY \cite{cospy}     & 63.9/70.2  & 71.6/83.9  & 58.5/55.8  & 69.6/83.4  & 65.7/79.7 & 65.7/76.1 \\ \hline
FerretNet \cite{ferretnet}          & \textcolor{blue}{98.0/100.0} & \textcolor{green}{92.6/98.5}  & 98.8/99.9  & 91.4/99.8  & 89.2/96.7 & 95.9/99.3 \\ \hline
DRIFT (Ours)   &  \textcolor{red}{98.6/100.0}  & \textcolor{blue}{94.3/99.3}  & \textcolor{blue}{99.1/100}  & \textcolor{green}{92.8/100}  & \textcolor{green}{91.2/97.8} & \textcolor{blue}{97.8/99.8} \\ \hline
\end{tabular}
\end{table}

\subsubsection{Testing Datasets.}
\label{sec:testing_data_supp}

We evaluate our method on the same benchmark suite adopted in FerretNet \cite{ferretnet} to ensure fair comparison. The evaluation benchmark contains real images and synthetic images generated by multiple state-of-the-art generative models, including both GAN-based and diffusion-based approaches. Specifically, we test on the following datasets:

\begin{itemize}
    \item \textbf{ForenSynths}:
    A forensic-oriented benchmark containing synthetic images generated by multiple GAN-based architectures: StyleGAN \cite{style}, BigGAN \cite{biggan}, CycleGAN \cite{cyclegan}, GauGAN \cite{gaunet}, and Deepfake \cite{depfake++}. Real images are sourced from six widely-used datasets: LSUN \cite{lsun}, ImageNet \cite{imagenet}, CelebA \cite{celeb}, CelebA-HQ \cite{celebHQ}, COCO \cite{coco}, and FaceForensics++ \cite{depfake++}.

    \item \textbf{Diffusion-6-Class (Diffusion-6cls)}:
    A diffusion-based benchmark comprising synthetic images generated by six distinct diffusion model variants, representing modern large-scale generative systems including DALL-E \cite{dalle}, Guided \cite{guided}, PNDM \cite{pndm}, VQ-Diffusion \cite{vqdiffusion}, Glide \cite{glide}, and LDM \cite{ldm}. Variants produced by Glide
and LDM with different parameter configurations are treated as separate categories. Each subset includes 1,000 synthetic and 1,000 real images.

    \item \textbf{Synthetic-Pop}:
    A collection of stylistically vivid synthetic images characterized by high-frequency textures and diverse semantic categories. This dataset captures latest  progress in high-resolution image generation, including Openjourney \cite{openjourney}, Proteus-0.3 \cite{proteus}, RealVisXL-4.0, SD-3.5-Medium \cite{scaling}, SDXL-Turbo \cite{sdxl}, and YiffyMix \cite{yiff}. Each subset includes 1,000 synthetic and 1,000 real images, with real images were drawn from COCO \cite{coco} and LAION-Aesthetics V2 (6.5+) \cite{laion} datasets.
\end{itemize}

\begin{table}[]
\caption{Accuracy and average precision comparisons  of different methods on Diffusion-6-cls test  dataset. The best, second-best and third-best results are highlighted in red, blue, and green colors, respectively.}
\label{table:2}
\begin{tabular}{|l|l|l|l|l|l|}
\hline
Dataset      & RIGID \cite{rigid}  & FatFormer \cite{fatformer} & SAFE \cite{improving_synthetic_detection} & FerretNet \cite{ferretnet} & DRIFT (Ours)       \\ \hline
Dall-E \cite{dalle}       & 43.5/48.0                  & \textcolor{red}{98.8/99.8}         & \textcolor{blue}{97.5/99.7}    & 91.4/98.2         &  \textcolor{green}{92.1/99.2}          \\ \hline
Guided \cite{guided}       &  58.0/94.9              & 76.1/92.0         & \textcolor{green}{82.4/95.8}    & \textcolor{blue}{92.1/98.6}         &  \textcolor{red}{92.9/98.9}          \\ \hline
PNDM \cite{pndm}         & 53.9/90.6                            & \textcolor{red}{99.3/100.0}        & 78.9/98.6    & \textcolor{blue}{96.9/100.0}        &  \textcolor{green}{94.3/100}          \\ \hline
VQ-Diffusion \cite{vqdiffusion} &  53.6/90.9                           & \textcolor{green}{98.0/100.0}       & \textcolor{red}{100.0/100.0}  & \textcolor{blue}{99.9/100.0}        &  97.0/100.0          \\ \hline
Glide-50-27 \cite{glide}  &   83.8/83.1                          & 94.7/99.4         & \textcolor{green}{96.6/99.2}    & \textcolor{red}{97.2/99.7}         &   \textcolor{blue}{96.8/99.3}         \\ \hline
Glide-100-10 \cite{glide} &  82.8/82.4                           & 94.2/99.2         & \textcolor{green}{97.3/99.4}    & \textcolor{red}{97.9/99.9}         &    \textcolor{blue}{97.4/100.0}        \\ \hline
Glide-100-27 \cite{glide} &  82.1/82.3                           & 94.4/99.1         & \textcolor{green}{95.8/98.9}    & \textcolor{red}{97.3/99.7}         &   \textcolor{blue}{96.4/99.4}         \\ \hline
LDM-100 \cite{ldm}      &   53.8/58.6                          & \textcolor{blue}{98.7/99.9}         & \textcolor{red}{98.8/100.0}   & 95/100.0        & \textcolor{green}{98.6/100.0} \\ \hline
LDM-200 \cite{ldm}      &   53.8/58.4                          & \textcolor{blue}{98.6/99.8 }        & 94.2/100.0   & \textcolor{red}{98.8/100.0}        &   \textcolor{green}{98.5/100.0}         \\ \hline

\end{tabular}
\end{table}

\begin{table}[]
\caption{Accuracy and average precision comparisons  of different methods on Synthetic-Pop test  dataset. The best, second-best and third-best results are highlighted in red, blue, and green colors, respectively.}
\label{table:3}
\begin{tabular}{|l|l|l|l|l|}
\hline
Methods           & Openjourney \cite{openjourney} & Proteus-0.3 \cite{proteus}  & SD-3.5-Medium \cite{sd35} & SDXL-Turbo \cite{sdxl} \\ \hline
FreqNet \cite{freqnet}   & 56.3/63.6   & 44.0/41.2      & 78.5/86.8     & 77.5/86.0  \\ \hline
NPR \cite{rethinking_upsampling}       & 78.8/83.5   & 68.6/69.3        & 80.4/84.1     & 78.2/82.9  \\ \hline
FatFormer \cite{fatformer} & 58.8/65.4   & 93.9/97.6       & 81.9/89.1     & 58.7/65.3  \\ \hline
SAFE \cite{improving_synthetic_detection}      & \textcolor{green}{94.7/99.3}   & \textcolor{red}{99.2/99.9}       & \textcolor{red}{98.1/99.7}     & \textcolor{green}{98.1/99.8}  \\ \hline
CO-SPY \cite{cospy}     & 92.4/97.6   & 88.8/93.0        & 80.9/87.8     & 79.9/88.3  \\ \hline
FerretNet \cite{ferretnet}        & \textcolor{red}{98.4/99.7}   & \textcolor{blue}{98.6/99.7}        & \textcolor{green}{97.2/99.6}     & \textcolor{red}{98.9/100.0} \\ \hline
DRIFT (Ours)              & \textcolor{blue}{97.4/99.1}   & \textcolor{green}{96.6/98.3}        & \textcolor{blue}{97.5/98.3}     & \textcolor{blue}{98.1/100.0} \\ \hline
\end{tabular}
\end{table}

\subsection{Main Results}

\subsubsection{Results on GAN-based generators (ForenSynths).}
On the ForenSynth benchmark, which includes GAN-based and Deepfake images, our method achieves consistently strong performance across all generator families as shown in Table~\ref{table:1}. In particular, the proposed approach achieves near-perfect detection for several generators such as StyleGAN and CycleGAN while maintaining strong performance on more challenging cases such as BigGAN and Deepfake. Compared with recent learning-based detectors and training-free robustness methods, our approach improves the overall mean performance and achieves a mean ACC/AP of approximately $97.8/99.8$. These results indicate that explicitly learning the invariance manifold of real images leads to a more stable detection signal than relying solely on pretrained feature geometry.

\subsubsection{Results on diffusion-based generators (Diffusion-6cls).}
We further evaluate in Table~\ref{table:2} the performance of various methods on the Diffusion-6cls benchmark containing multiple diffusion-based generators. Diffusion-generated images are known to be significantly more challenging to detect due to their high visual fidelity and reduced artifacts. Despite these challenges, our method achieves consistently high detection accuracy across all diffusion variants, often approaching perfect ACC and AP values. Compared with training-free detectors such as RIGID and reconstruction-based approaches such as FIRE, our method provides more stable performance across different diffusion sampling strategies. This demonstrates that the learned invariance manifold captures structural properties of real images that generalize beyond specific generator architectures.

\subsubsection{Results on text-to-image generators (Synthetic-Pop).}
The Synthetic-Pop dataset contains recent high-resolution text-to-image generators. These generators produce images with strong semantic realism and high-frequency details, making detection particularly difficult. Our method achieves strong results across all generators, with ACC values typically above $96\%$ and near-perfect AP scores as shown in Table~\ref{table:3}. Compared with previous detectors, our approach remains stable across stylistically diverse generators, highlighting its robustness to distribution shifts in generative models.

\begin{figure}[!]
\centering
\begin{subfigure}{0.45\textwidth}
\centering
\includegraphics[width=\textwidth]{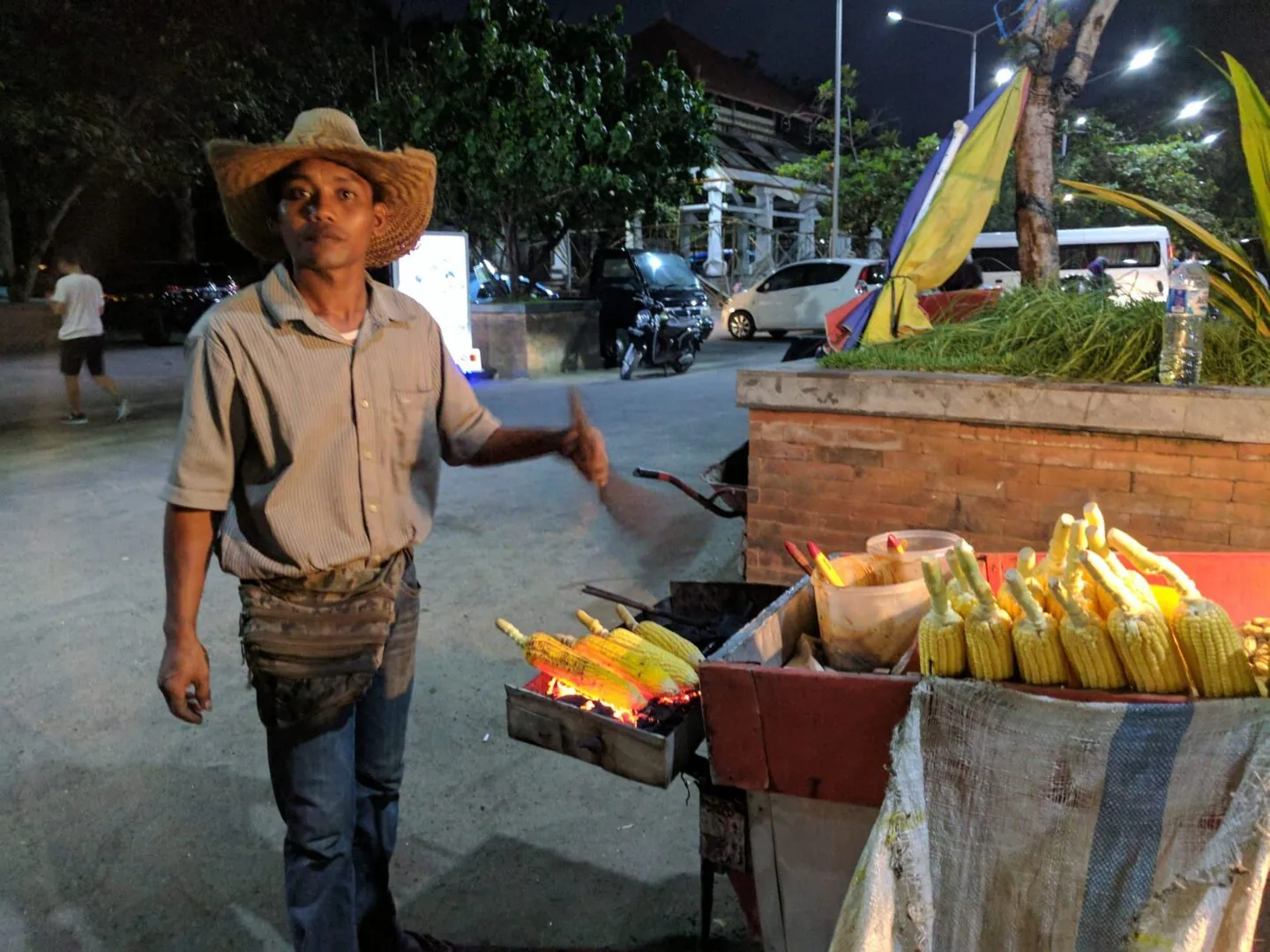}
\end{subfigure}
\hfill
\begin{subfigure}{0.45\textwidth}
\centering
\includegraphics[width=\textwidth]{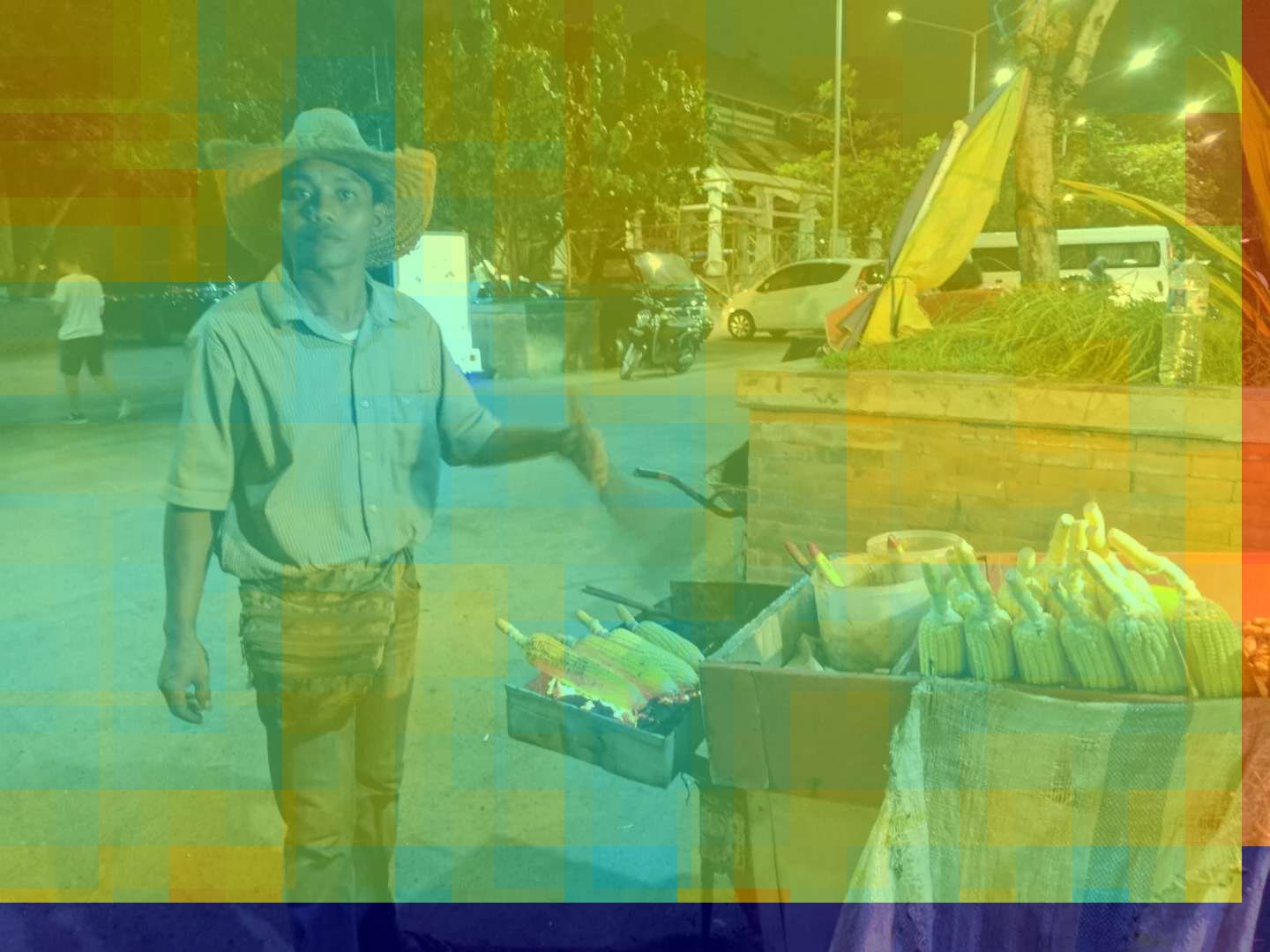}
\end{subfigure}
\hfill
\begin{subfigure}{0.45\textwidth}
\centering
\includegraphics[width=\textwidth]{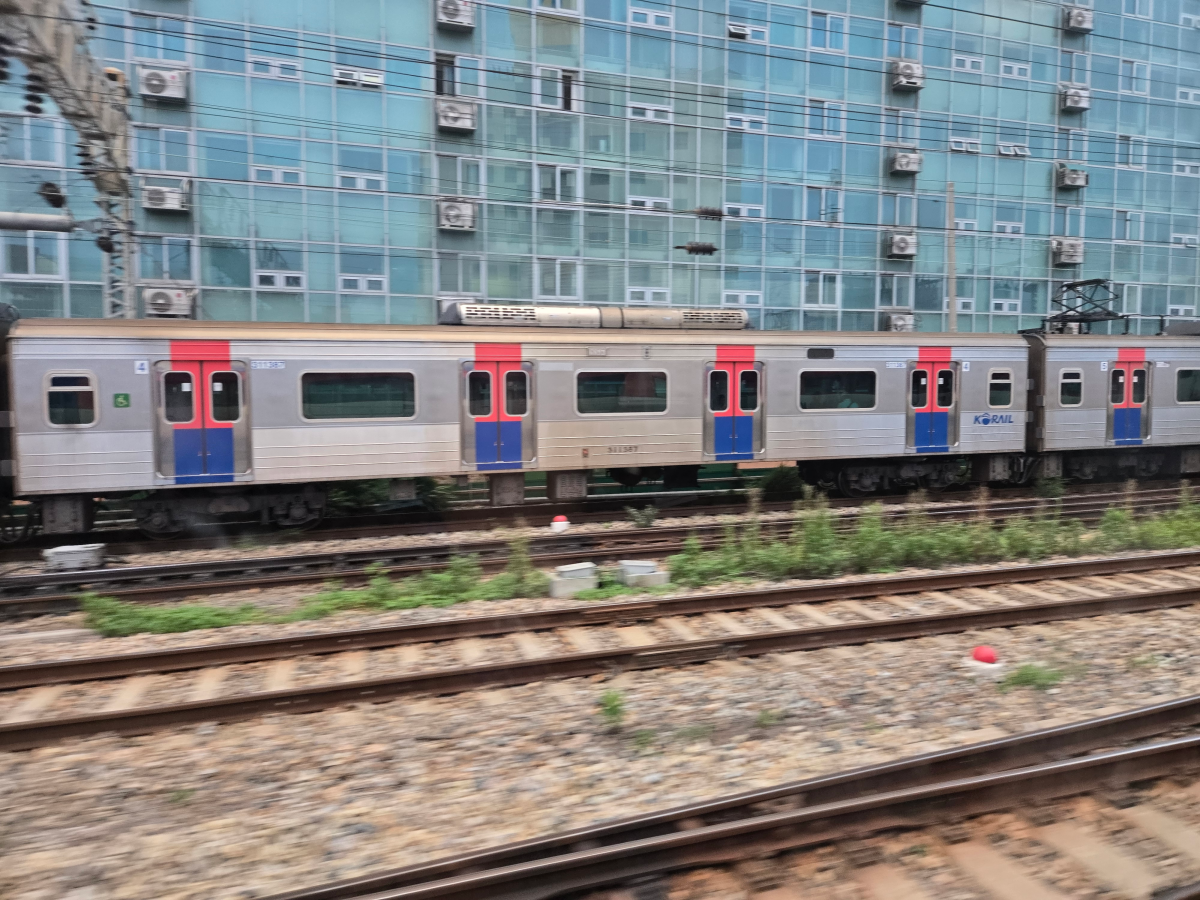}
\end{subfigure}
\hfill
\begin{subfigure}{0.45\textwidth}
\centering
\includegraphics[width=\textwidth]{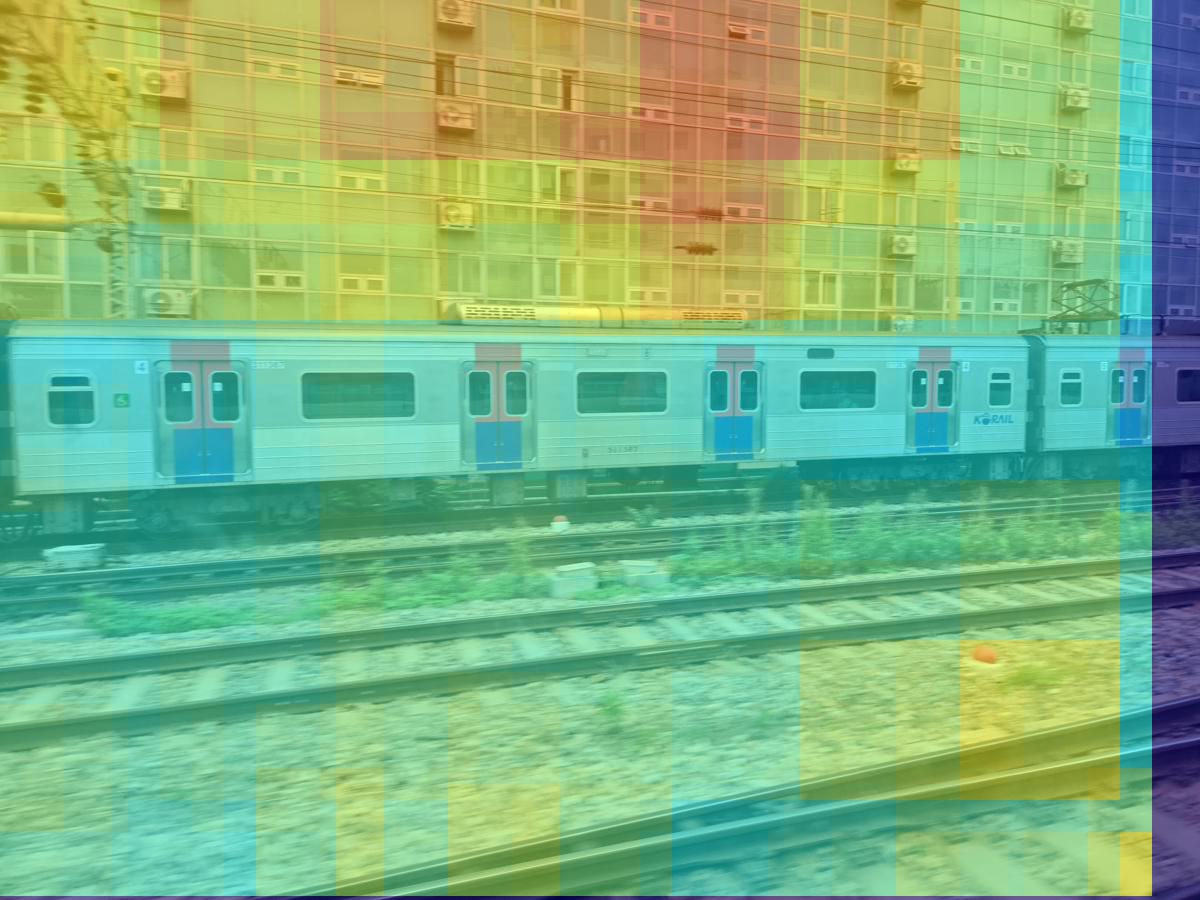}
\end{subfigure}
\hfill
\begin{subfigure}{0.45\textwidth}
\centering
\includegraphics[width=\textwidth]{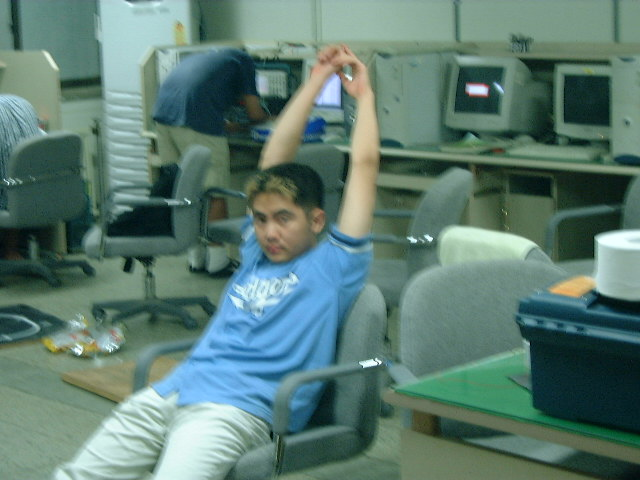}
\caption{Real image.}
\end{subfigure}
\hfill
\begin{subfigure}{0.45\textwidth}
\centering
\includegraphics[width=\textwidth]{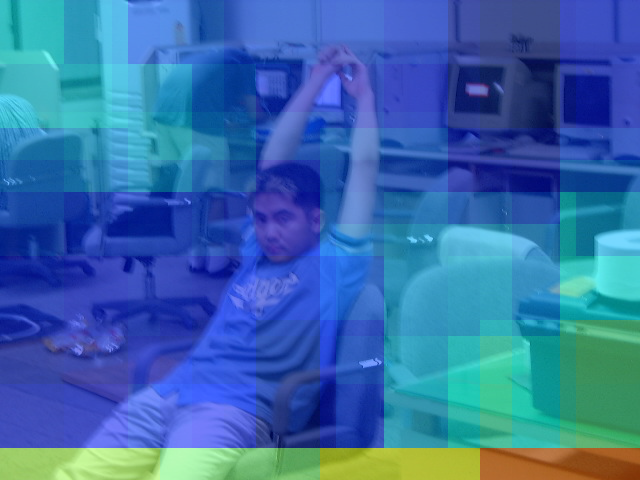}
\caption{Real image: Detection heatmap.}
\end{subfigure}
\caption{Visualization of robust-fragile drift violation. Real images maintain drift consistency.}
\label{fig:real1}
\end{figure}

\begin{figure}[!]
\centering
\begin{subfigure}{0.45\textwidth}
\centering
\includegraphics[width=\textwidth]{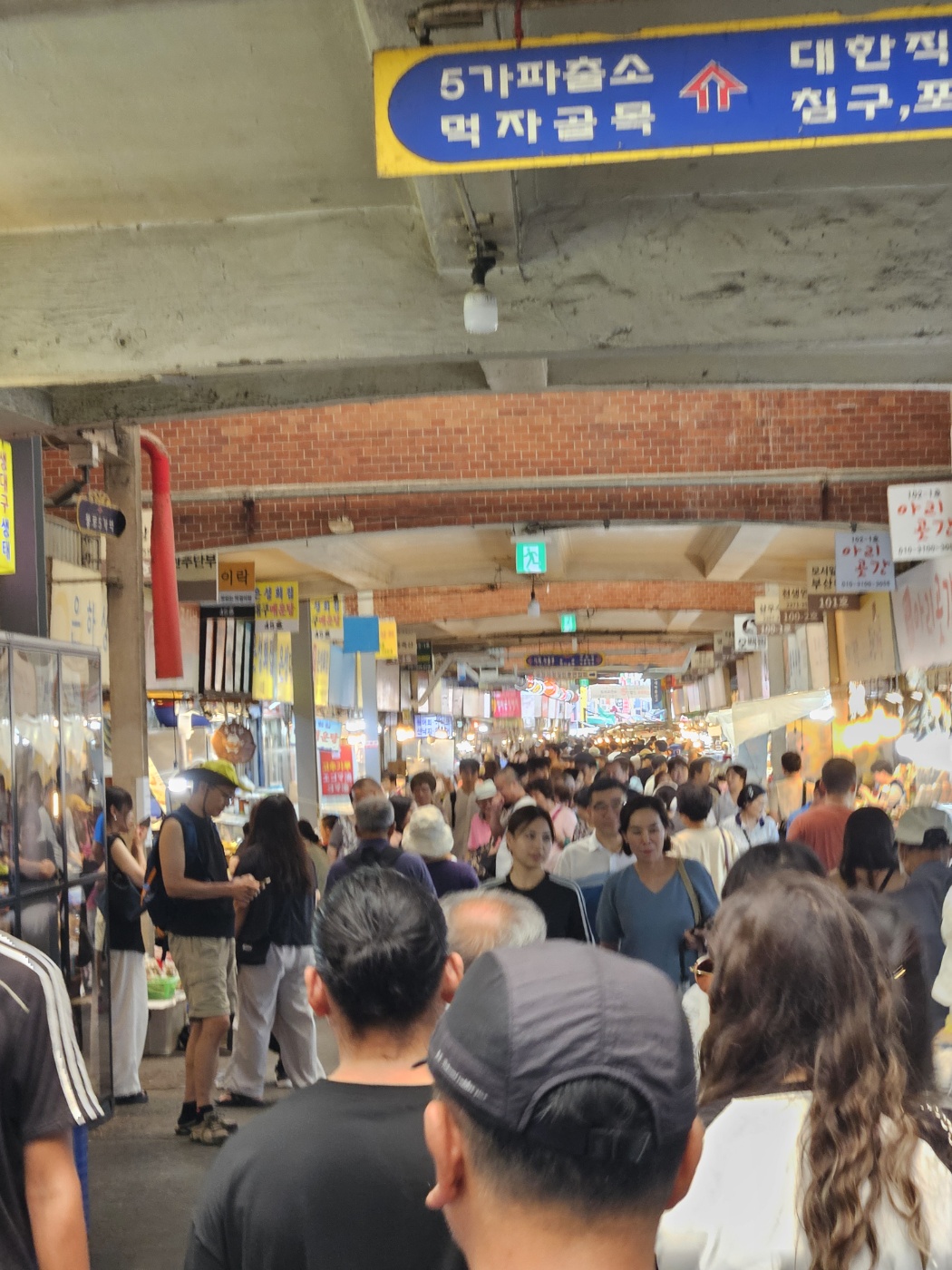}
\end{subfigure}
\hfill
\begin{subfigure}{0.45\textwidth}
\centering
\includegraphics[width=\textwidth]{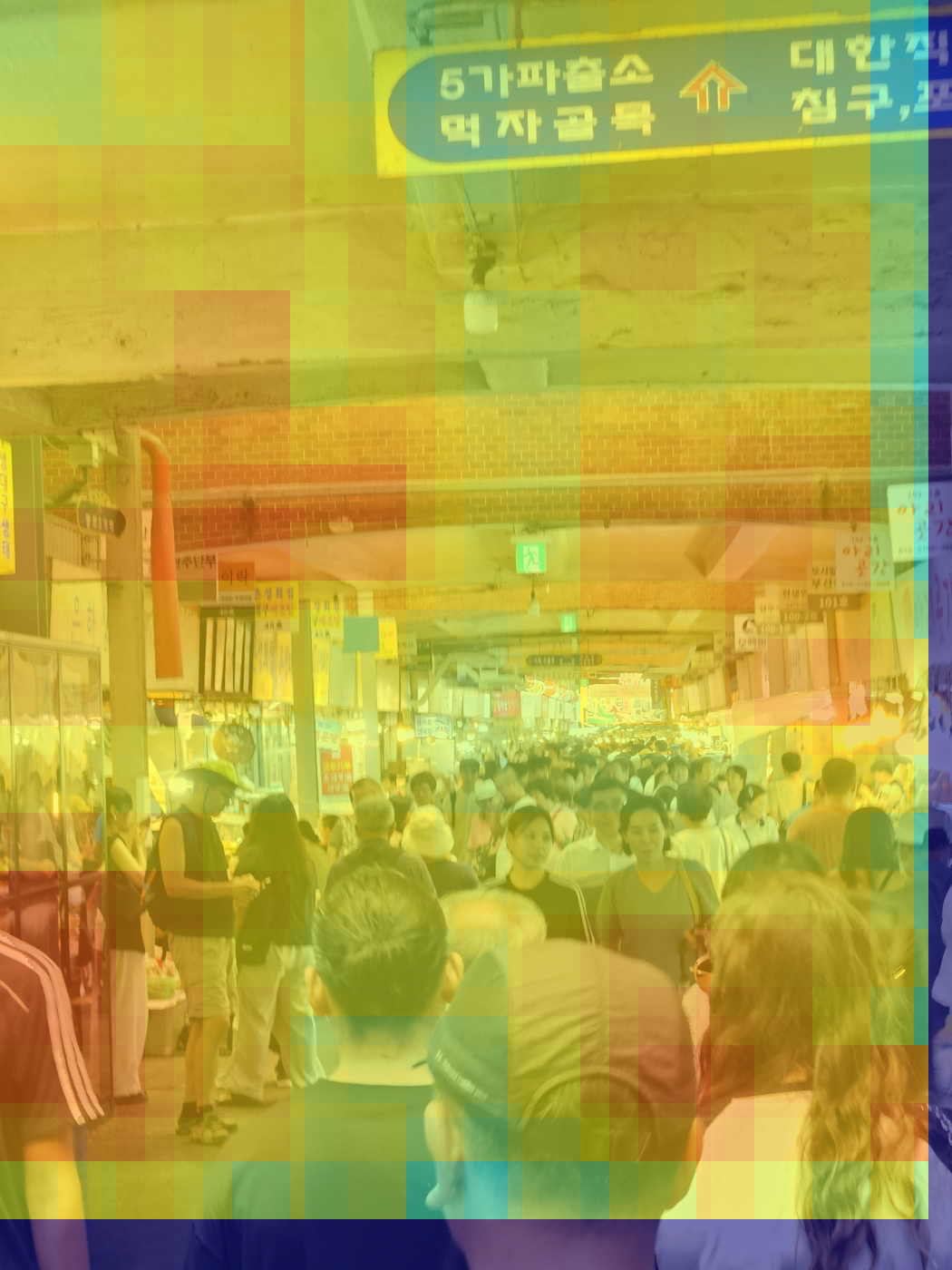}
\end{subfigure}
\hfill
\begin{subfigure}{0.45\textwidth}
\centering
\includegraphics[width=\textwidth]{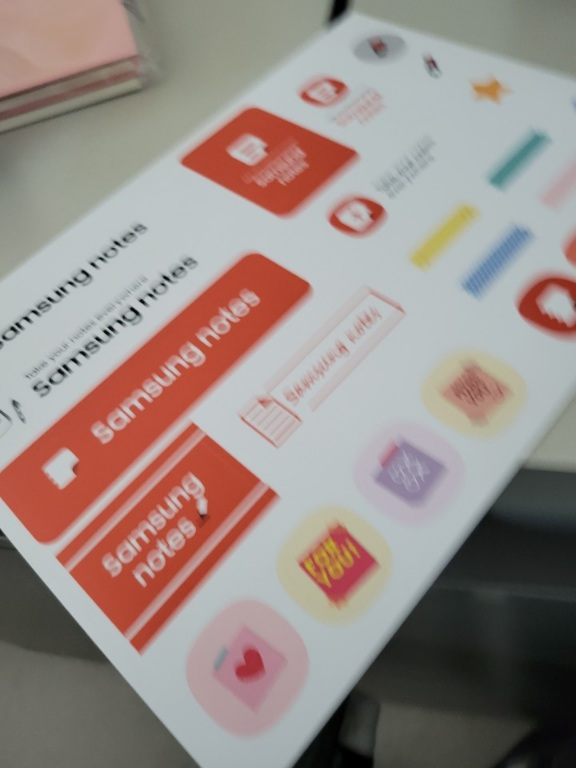}
\caption{Real image.}
\end{subfigure}
\hfill
\begin{subfigure}{0.45\textwidth}
\centering
\includegraphics[width=\textwidth]{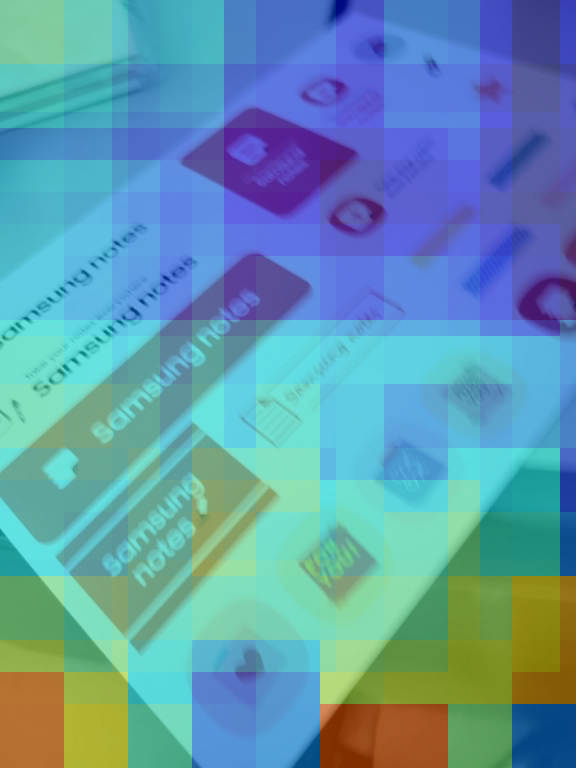}
\caption{Real: Detection heatmap.}
\end{subfigure}
\caption{Visualization of robust-fragile drift violation. Real images maintain drift consistency.}
\label{fig:real2}
\end{figure}

\subsection{Qualitative Analysis}

Figs.~\ref{fig:real1} and \ref{fig:real2} provide qualitative evidence that real images largely preserve robust-fragile drift consistency. Although the detector produces spatially varying responses, the corresponding heatmaps remain relatively smooth and moderate in intensity, without large contiguous regions of saturated activation. This behavior is consistent with our hypothesis that, for real images, both robust and fragile transformations induce representation changes that remain locally compatible with the natural-image manifold. In other words, even when perturbations alter texture, blur level, or local photometric statistics, the induced feature drift does not strongly violate the learned invariance structure. The responses observed in these figures therefore reflect normal acquisition variability rather than synthetic inconsistency.

Fig.~\ref{fig:real1} is particularly informative because it includes examples with diverse scene structure, including people, man-made objects, and motion-corrupted content. Despite these variations, the detector does not collapse to object-specific cues; instead, it produces broadly distributed and comparatively low-confidence responses. This suggests that the proposed criterion is not simply reacting to semantics, foreground saliency, or scene complexity. Rather, it is measuring whether the image remains stable under the contrast between robust and fragile perturbation pathways. Similarly, Fig.~\ref{fig:real2} shows that even in crowded scenes and close-up content with strong perspective or blur, the detector still avoids the highly concentrated high-response patterns that are typical of synthetic samples. These examples support the claim that real images occupy regions where the robust and fragile feature trajectories remain mutually consistent.

\begin{figure}[!]
\centering
\begin{subfigure}{0.45\textwidth}
\centering
\includegraphics[width=\textwidth]{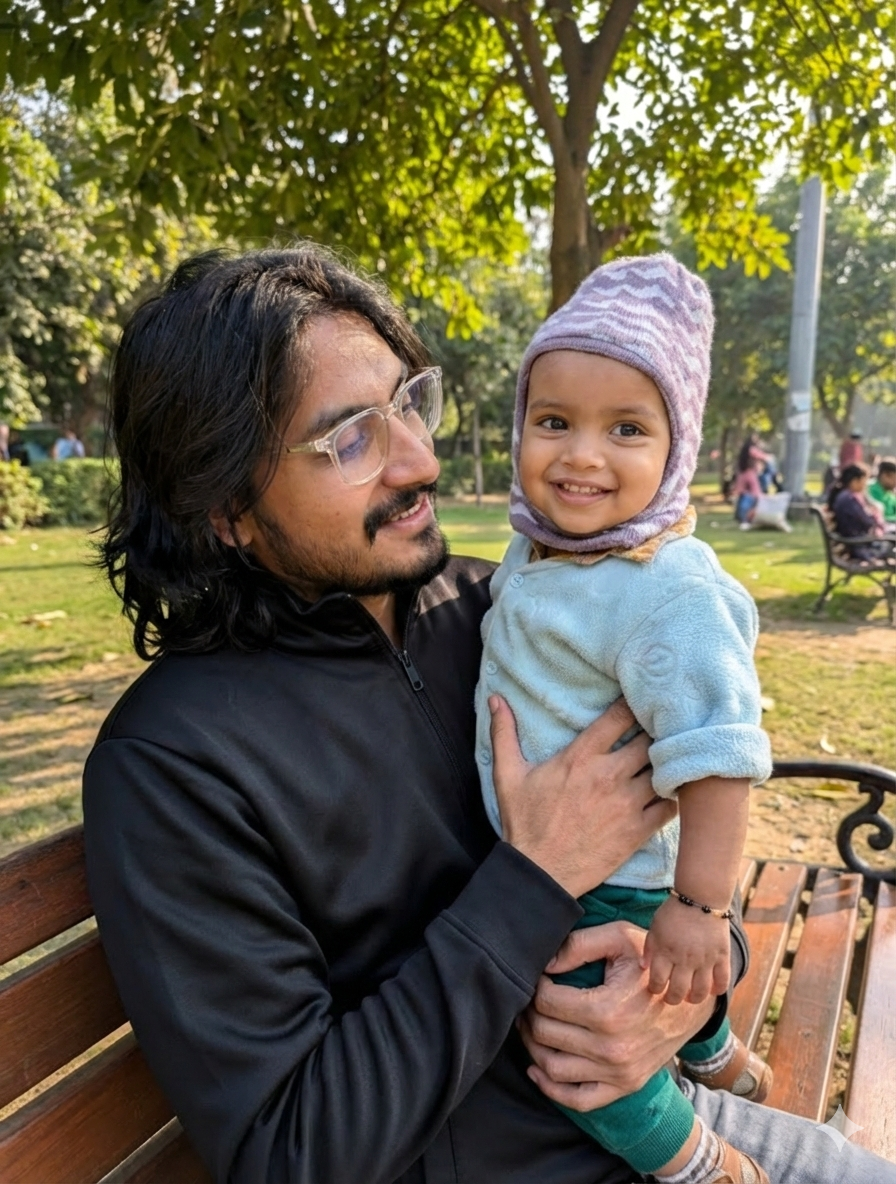}
\end{subfigure}
\hfill
\begin{subfigure}{0.45\textwidth}
\centering
\includegraphics[width=\textwidth]{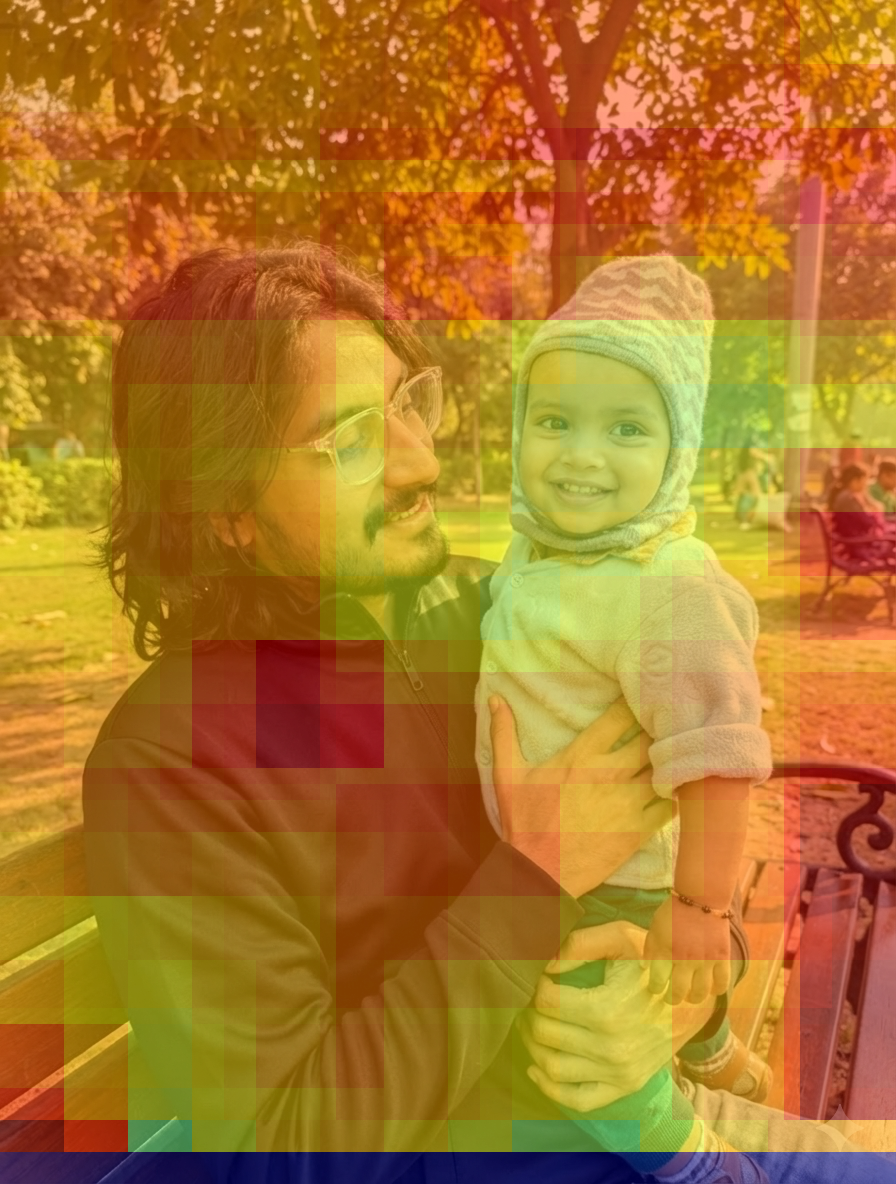}
\end{subfigure}
\hfill
\begin{subfigure}{0.45\textwidth}
\centering
\includegraphics[width=\textwidth]{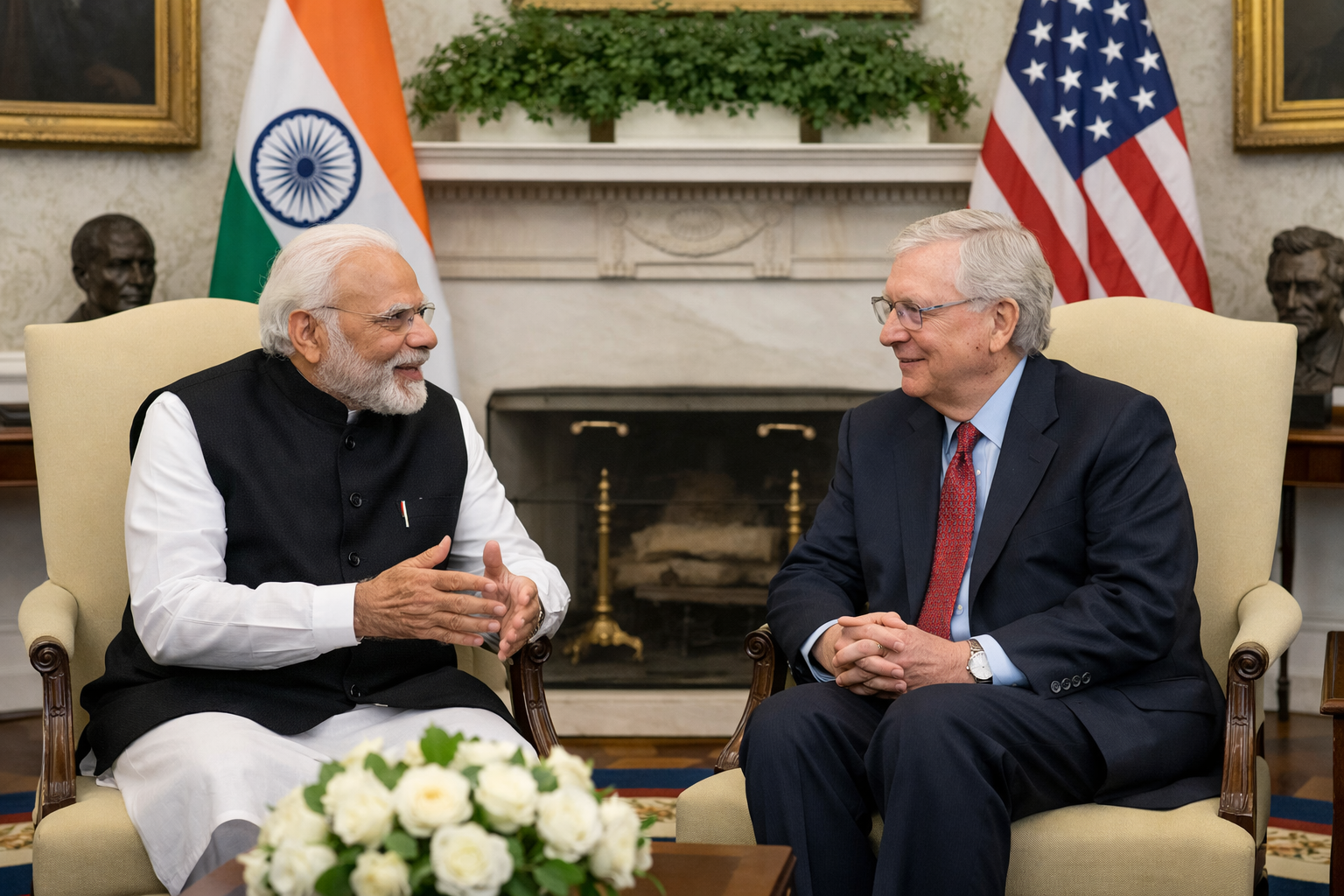}
\end{subfigure}
\hfill
\begin{subfigure}{0.45\textwidth}
\centering
\includegraphics[width=\textwidth]{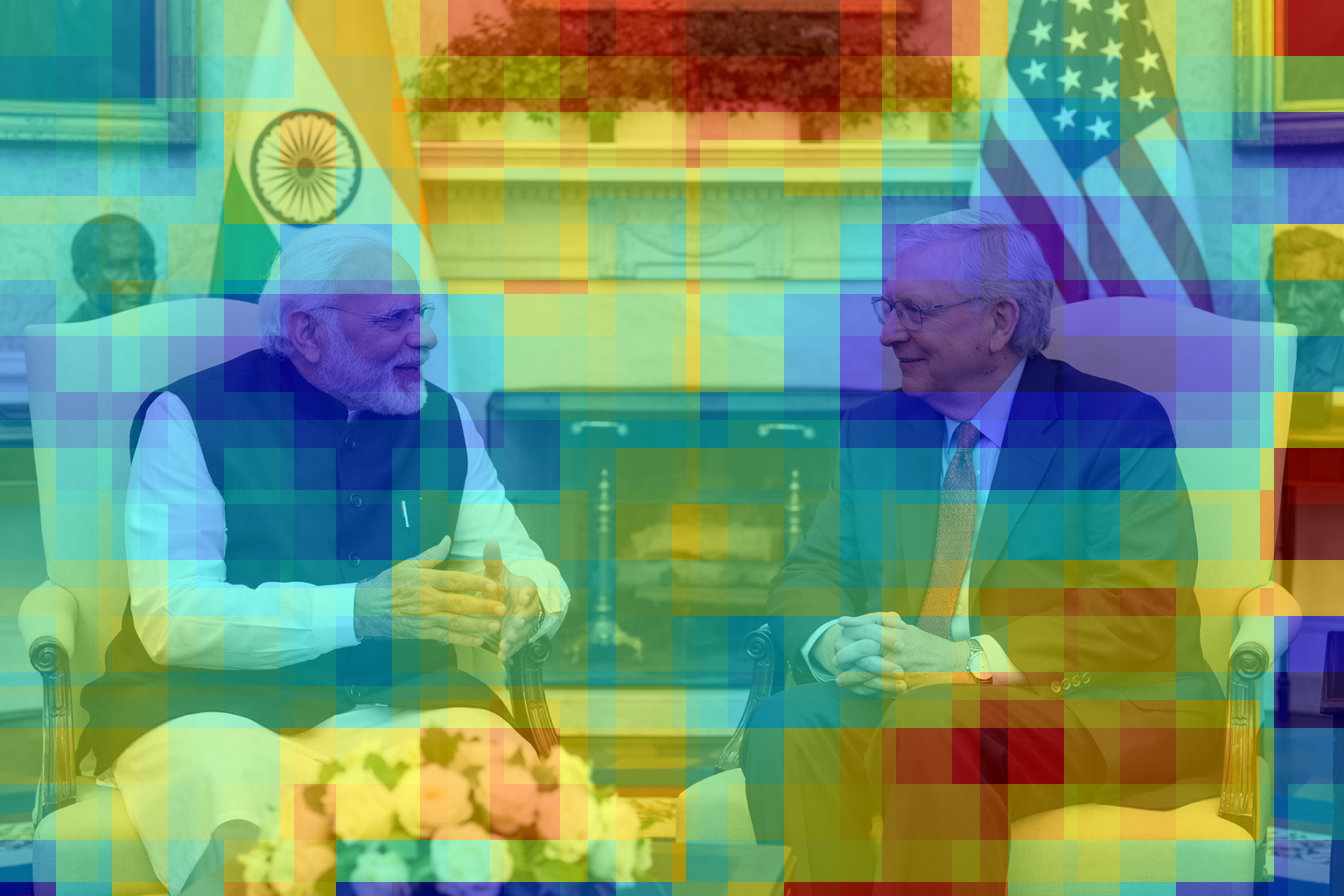}
\end{subfigure}
\hfill
\begin{subfigure}{0.45\textwidth}
\centering
\includegraphics[width=\textwidth]{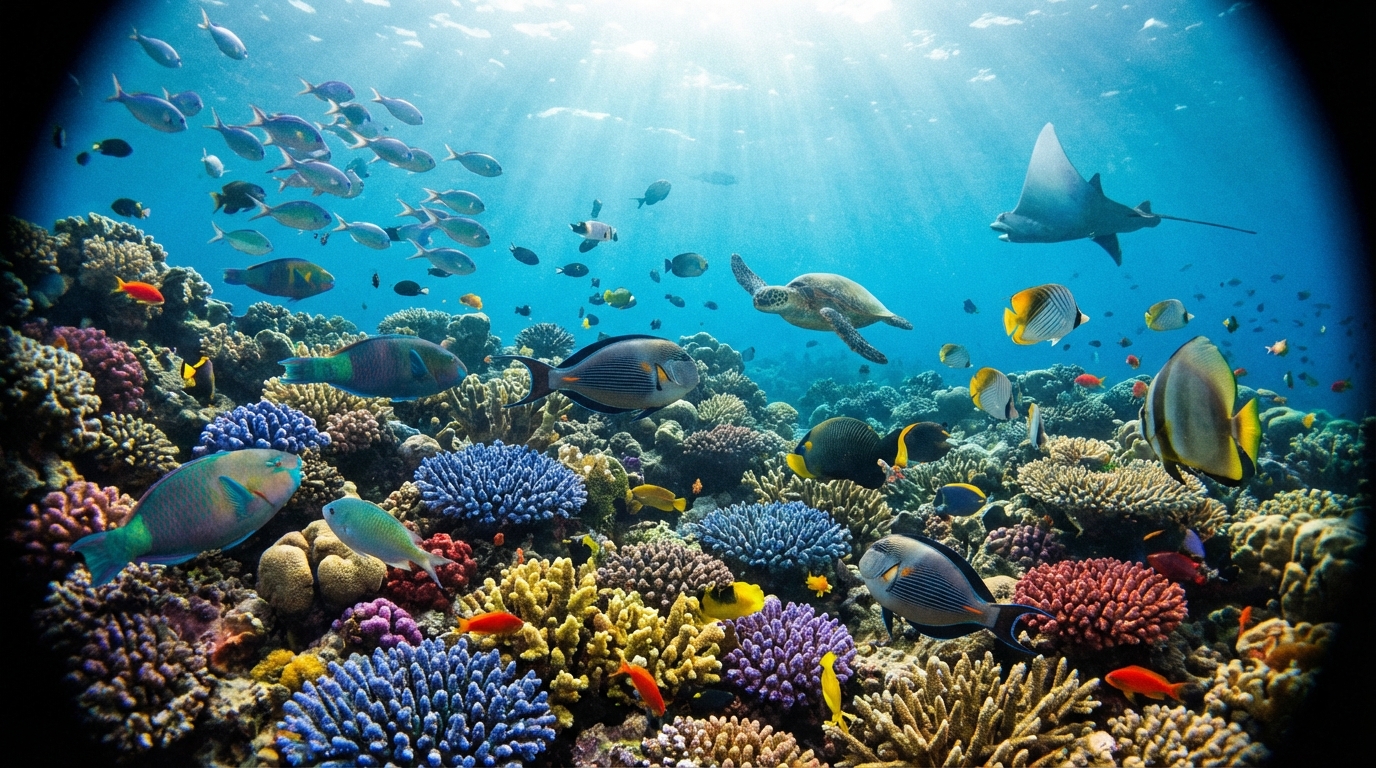}
\caption{AI-generated image.}
\end{subfigure}
\hfill
\begin{subfigure}{0.45\textwidth}
\centering
\includegraphics[width=\textwidth]{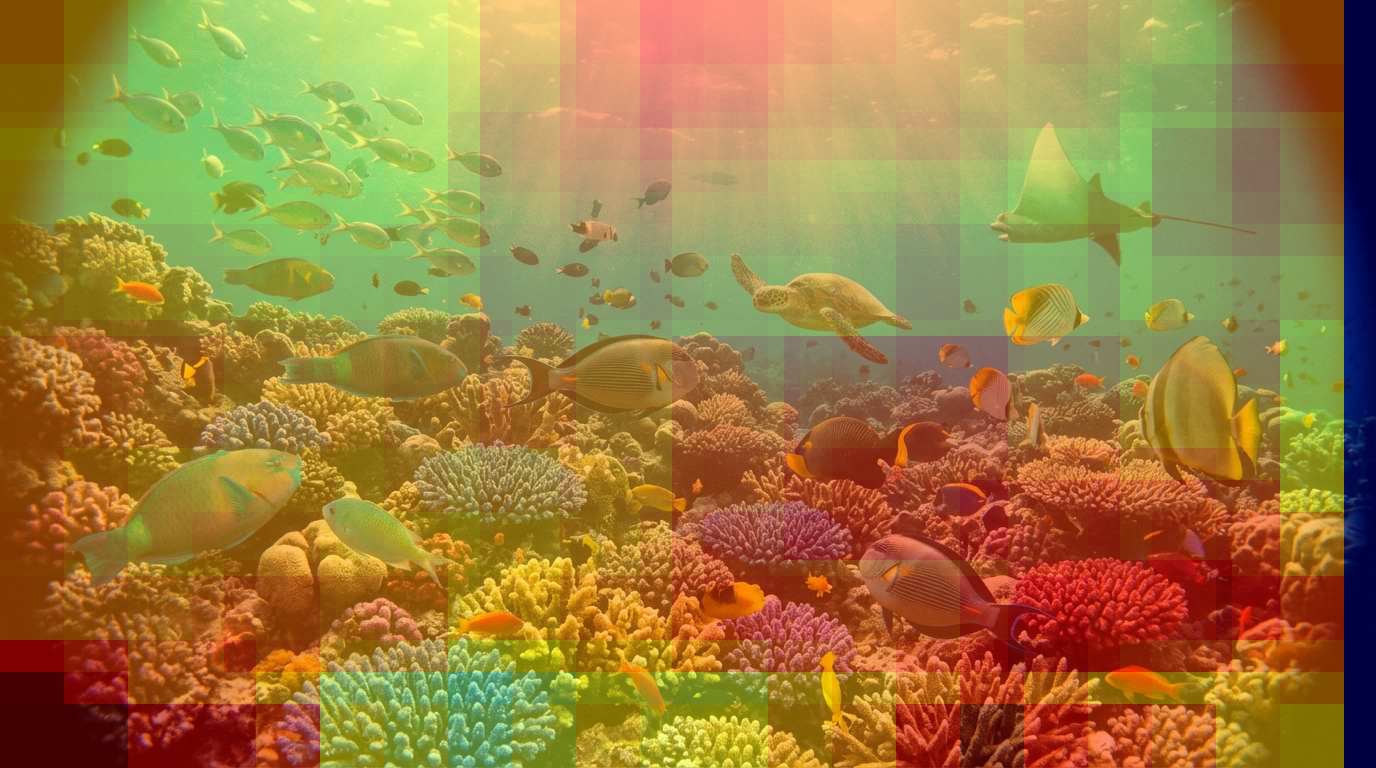}
\caption{AI-generated image: Detection heatmap.}
\end{subfigure}
\caption{Visualization of robust-fragile drift violation. AI-generated images show strong drift violations, producing high-response heatmaps.}
\label{fig:synth1}
\end{figure}

\begin{figure}[!]
\centering
\begin{subfigure}{0.45\textwidth}
\centering
\includegraphics[width=\textwidth]{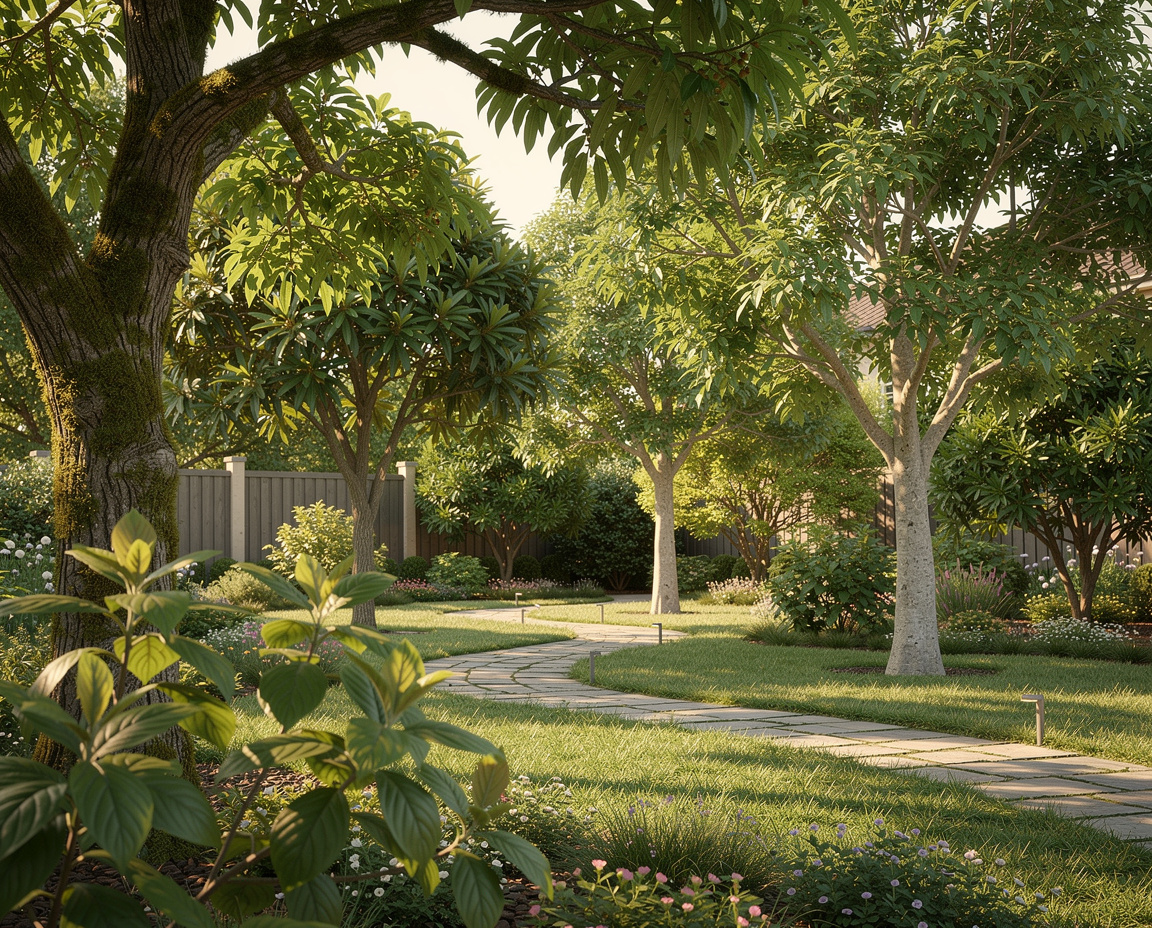}
\end{subfigure}
\hfill
\begin{subfigure}{0.45\textwidth}
\centering
\includegraphics[width=\textwidth]{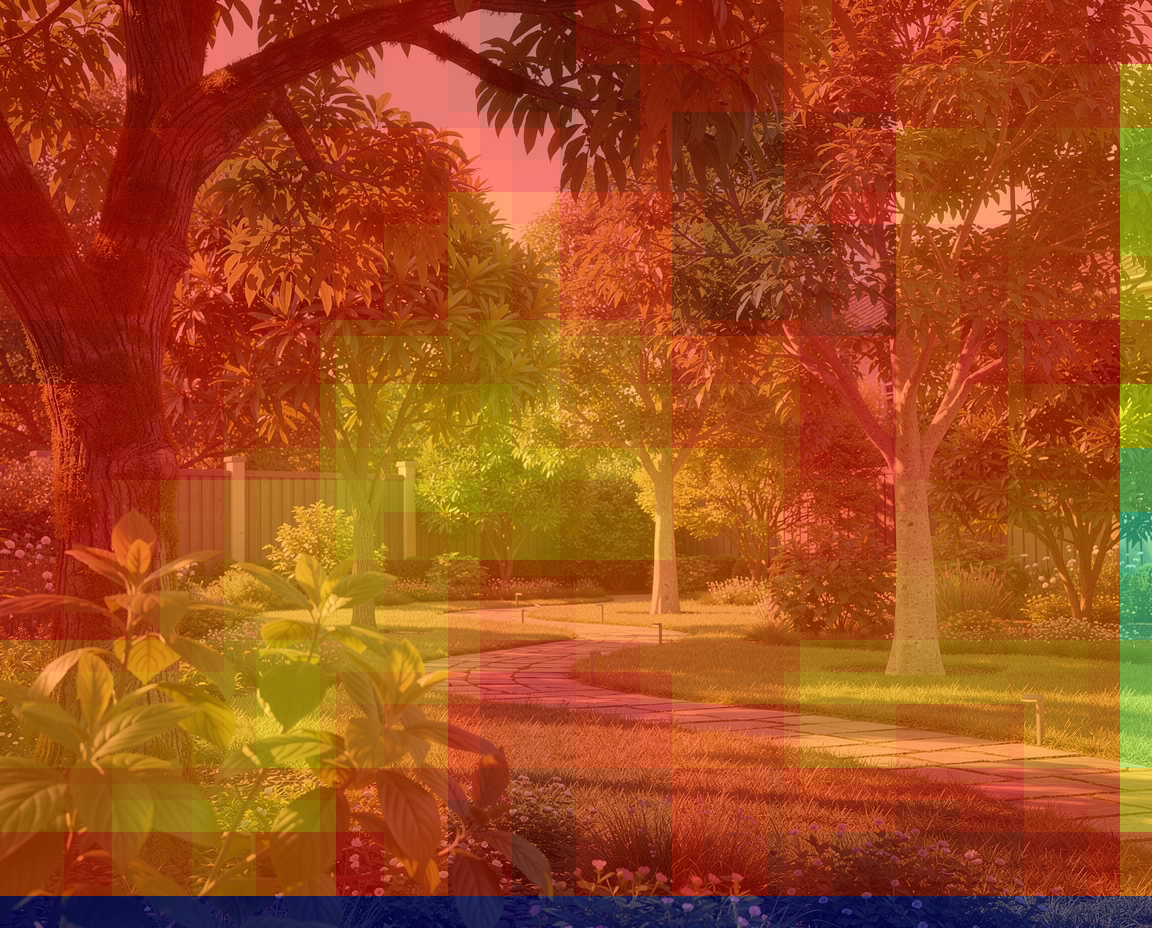}
\end{subfigure}
\hfill
\begin{subfigure}{0.45\textwidth}
\centering
\includegraphics[width=\textwidth]{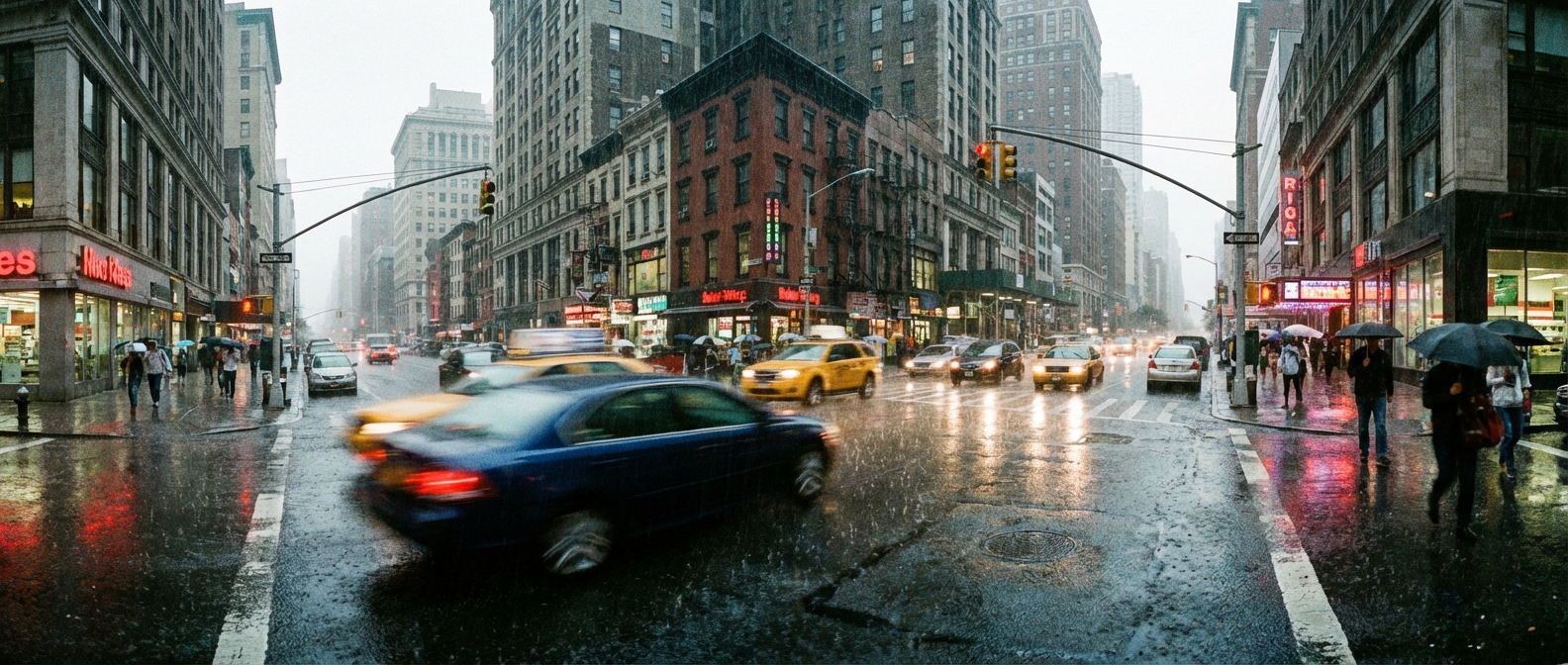}
\end{subfigure}
\hfill
\begin{subfigure}{0.45\textwidth}
\centering
\includegraphics[width=\textwidth]{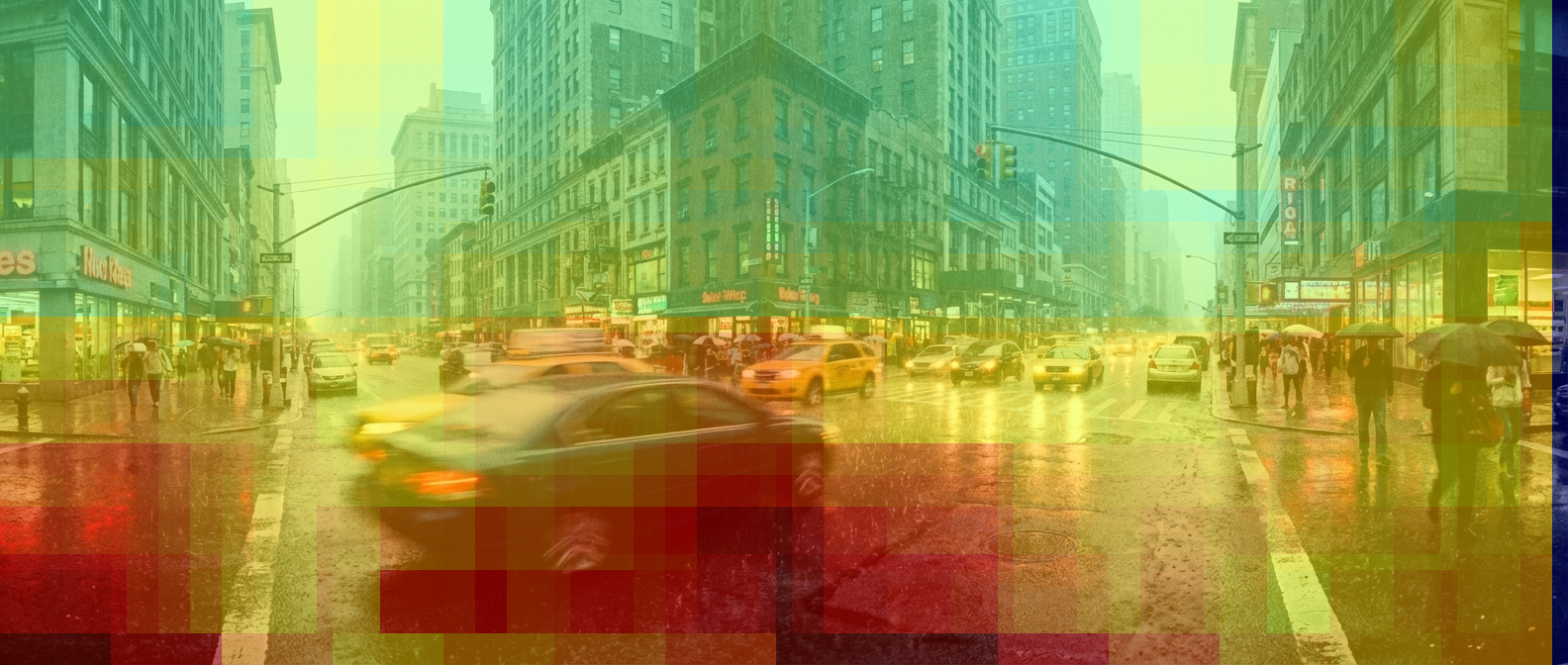}
\end{subfigure}
\hfill
\begin{subfigure}{0.45\textwidth}
\centering
\includegraphics[width=\textwidth]{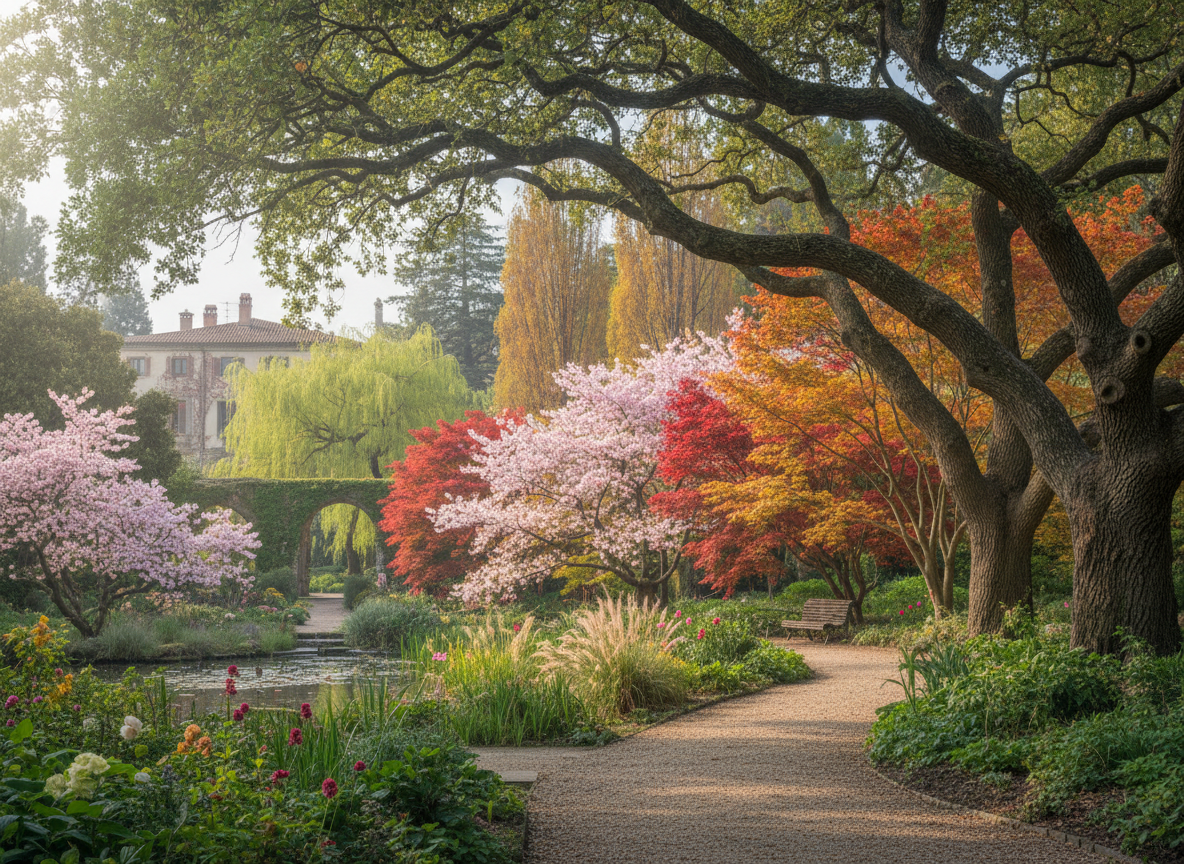}
\end{subfigure}
\hfill
\begin{subfigure}{0.45\textwidth}
\centering
\includegraphics[width=\textwidth]{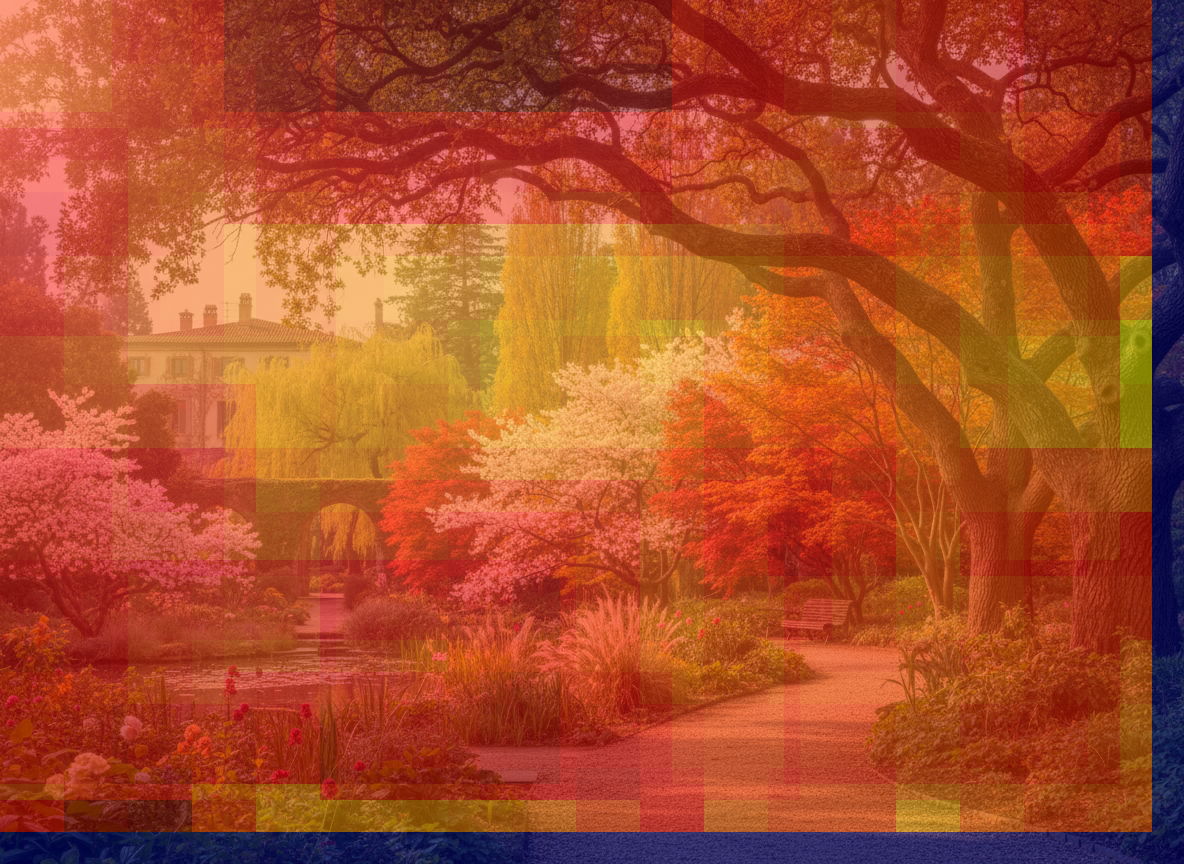}
\end{subfigure}
\hfill
\begin{subfigure}{0.45\textwidth}
\centering
\includegraphics[width=\textwidth]{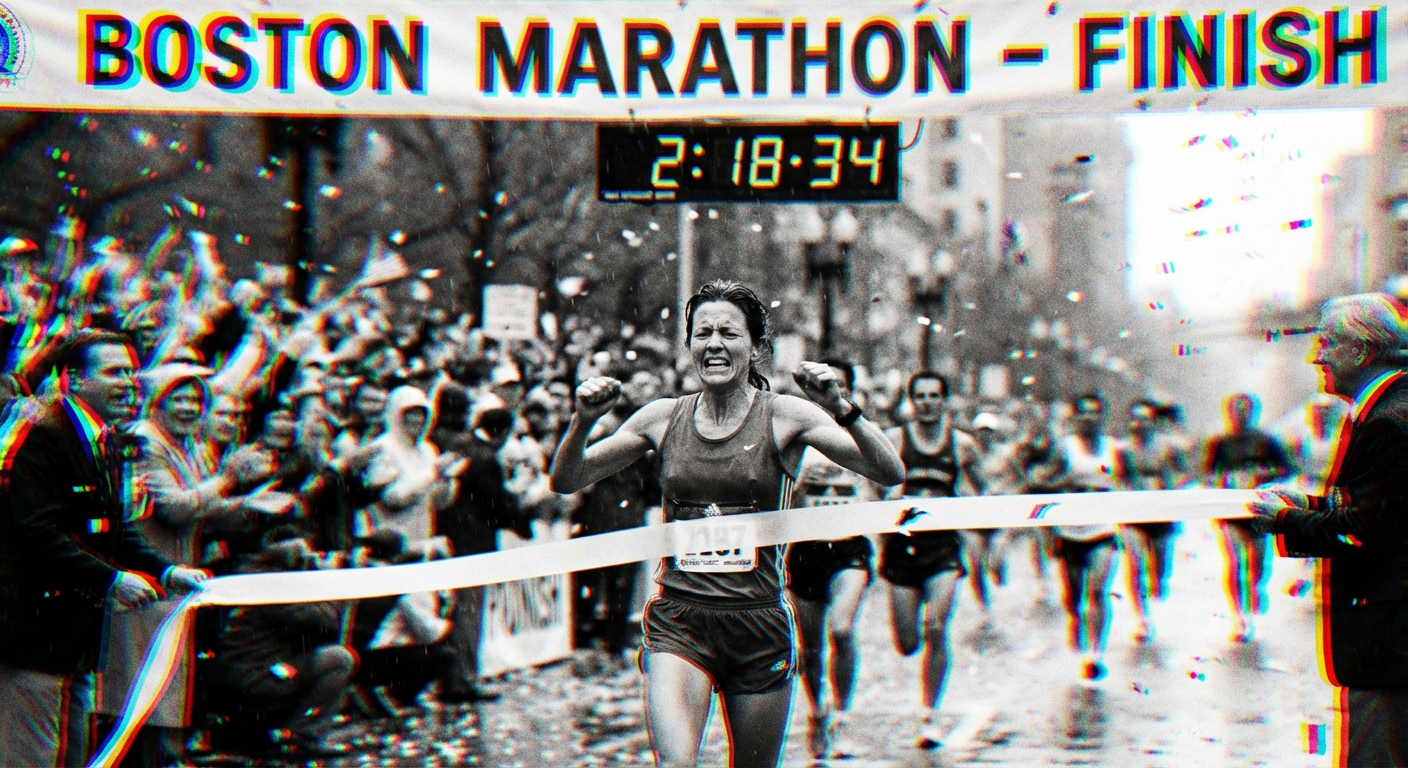}
\caption{AI-generated image.}
\end{subfigure}
\hfill
\begin{subfigure}{0.45\textwidth}
\centering
\includegraphics[width=\textwidth]{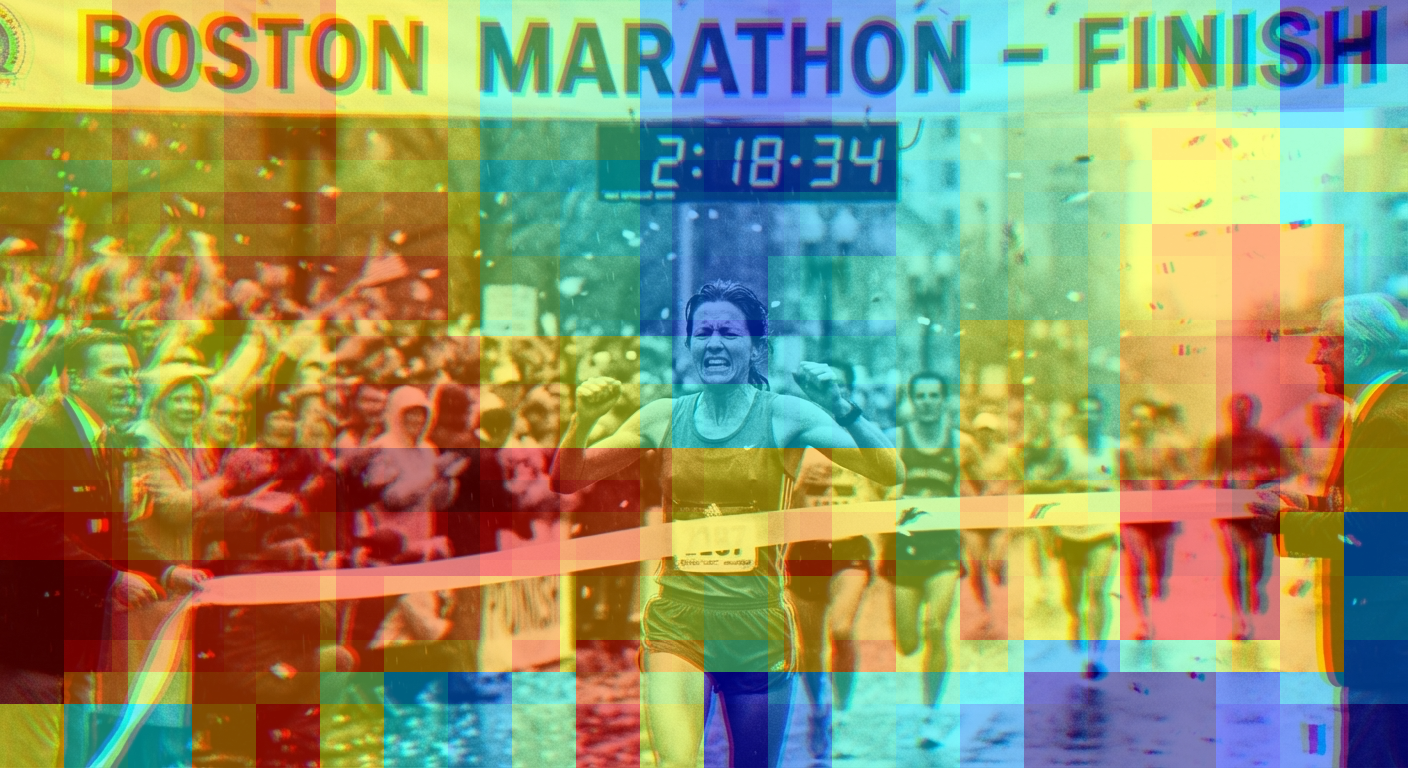}
\caption{AI-generated image: Detection heatmap.}
\end{subfigure}
\caption{Visualization of robust-fragile drift violation. AI-generated images show strong drift violations, producing high-response heatmaps.}
\label{fig:synth2}
\end{figure}

In contrast, Figs.~\ref{fig:synth1} and \ref{fig:synth2} show that AI-generated images produce substantially stronger and more spatially extensive activations. The heatmaps are noticeably more intense and often cover semantically meaningful regions such as faces, text-bearing areas, structural boundaries, or highly textured content. This indicates that the representations extracted from synthetic images are much less stable under the discrepancy between robust and fragile transformations. From the manifold perspective introduced in the supplementary theory section, these responses are consistent with larger off-manifold deviations: fragile transformations expose directions in feature space that are not well aligned with the local geometry of real images, causing a pronounced drift violation signal.

Another notable observation in Figs.~\ref{fig:synth1} and \ref{fig:synth2} is that the detector responds not only to obviously unrealistic artifacts, but also to globally coherent synthetic images that may appear visually plausible at first glance. For example, even when object layout, illumination, and color composition are convincing, the heatmaps still reveal broad inconsistency patterns. This suggests that the proposed method captures a deeper structural property than superficial artifact detection. Instead of depending on a single local anomaly, it aggregates evidence of instability across multiple regions and scales, which is especially important for modern generative models whose outputs are often photorealistic in isolated patches but inconsistent in their transformation behavior.

Taken together, these supplementary visualizations reinforce the central claim of the paper: real images tend to preserve drift consistency under the proposed transformation families, whereas AI-generated images exhibit stronger invariance violations that manifest as high-response heatmaps. The qualitative examples therefore complement the quantitative results by showing that the detector's decision is spatially interpretable and aligned with the intended mechanism. They also indicate that the method generalizes across diverse image content, including portraits, natural scenes, urban environments, text-rich regions, and crowded real-world photographs, further supporting the robustness of the proposed drift-based detection principle.

